\documentclass[10pt,a4paper, oneside, final]{article}
\usepackage{epsfig,subfigure,float}

\begin{document}

\title{Diffusion of Information in Robot Swarms}

\author{Serge Kernbach \\
{\small Institute of Parallel and Distributed Systems, University of Stuttgart} \\
{\small Universit{\"a}tsstr.~38, D-70569 Stuttgart, Germany}\\
{\small \it Serge.Kernbach@ipvs.uni-stuttgart.de}
}
\date{}
\maketitle

\begin{abstract}
This work is devoted to communication approaches, which spread information in robot swarms. These mechanisms are useful for large-scale systems and also for such cases when a limited communication equipment does not allow routing of information packages. We focus on two approaches such as virtual fields and epidemic algorithms, discuss several aspects of hardware implementation and demonstrate experiments performed with microrobots "Jasmine".
\end{abstract}

\section{Introduction}
\label{sec_data_flow}

Swarm robotics differs from other fields of robotic systems in several essential points such as inaccessibility of global coordinates, global perception and global communication~\cite{Kernbach11-HCR}, \cite{Siciliano08}. These issues impact the utilized mechanisms of coordination, perception and control~\cite{KornienkoS05d}. Communication is one of the central mechanisms for collective systems, e.g.~\cite{Fukuda91}, \cite{Liu2010}, \cite{Marocco06}, it provides:
\begin{itemize}
    \item "awareness" for all robots in a swarm about events relevant for the whole swarm, e.g. availability of energy.
    \item macroscopic (non-local) coordination and collective decision making for the cases of global and feedback connectivity~\cite{2011arXiv1109.4221K}.
    \item local coordination for executing cooperative activities such as assembling, objects handling or collective perception~\cite{Jimenez05}.
\end{itemize}

Since robots possess a limited communication radius and move, they build a peer-to-peer communication network with a quickly changing structure, e.g.~\cite{Akyildiz04}. Within this network robots exchange small pieces of information: values of variables, some coded information about intensities~\cite{Kornienko_S04}, symbolically coded shapes for collective perception~\cite{Kornienko_S05a} and others. There are several basic mechanisms that allow propagating this information among robots in such networks:
\begin{enumerate}

\item Information exchange without direct communication. This approach can be divided into two parts: using physical medium for saving messages, such as electro-magnetic markers (or RFID tags) on objects, and using activities for indirect communication, e.g. if an activity "A" is finished, it is a signal for starting an activity "B". Situation awareness is automatically incorporated into a message. This type of information exchange is also known as stigmergy.

\item Direct communication between neighbors without routing of messages. In this simple approach all robots are expected to be aware of a context of messages. Peer-to-peer exchange of information serves for coordinating activities between local neighbors.

\item Direct communication with routing of messages. It is usually a package-based approach, widely used in distributed communication systems. Each package includes, at least, IDs of sender and receiver, as well as the information content. Situation awareness can be directly included into the content of messages.

\item Pheromone-based and epidemic approaches that are known from biological systems. Type of pheromone as well as secondary signal's information (intensity, direction) point to the context of messages.
\end{enumerate}

Information exchange through stigmergy~\cite{Bonabeau99}, and a direct communication is effective when a small group of robots performs homogeneous activities with well defined roles. We denote these mechanisms as a local swarm communication because they do not provide a propagation of information.

Direct communication between neighbors with routing of messages allows a global information transfer in a swarm~\cite{Kornienko_S05d}. Implementing these mechanisms, we encounter several problems related to identification of a spatio-temporal origin of messages (known also as context information~\cite{Ebeling98}) and to the management of messages. Statio-temporal context is used in different navigation and collaboration algorithms~\cite{Kornienko_S05b}. If a robot receives some message from neighbors, it does not know a spatial position of the initial sender and can only follow the propagation way. Since robots move, the propagation way, represented by e.g. a sequence of stored IDs, does not reflect neither spatial nor temporal state of the system. Taking into account also a high computational effort, the approach based on routing of packages is not useful in many swarm applications. If the routing is not used, robots only exchange information between local neighbors.

In this work we focus on a global communication without routing of packages. Attention is paid, in particular, on diffusion of information by means of virtual pheromones, epidemic algorithms~\cite{ma2009} and their embodiment~\cite{Kornienko_S05e}. Since these mechanisms spread information globally, we can employ them in different coordination and navigation techniques. In Sec.~\ref{sec:densityRadius} we give overview of these approaches, in Sec.~\ref{sec:hardware} introduce corresponding hardware and in Sec.~\ref{sec:Experiment} discuss preformed experiments. Finally, Sec.~\ref{sec:conclusion} concludes this paper.

\section{Swarm Communication, Virtual Fields and Epidemic Algorithms}
\label{sec:densityRadius}

State of the art approaches differentiate three main levels of communication: physical signal transmission, communication protocols and information structures that require communication. In swarm-based systems there exists an additional behavioral level related to creating and supporting dynamical communication networks.

\textbf{1.} On the level of physical signal transmission the communication deals with a choice of modulation/transmission approaches. There are different available solutions for IR-based signal transmission: single impulses, pulse code modulation (PCM) as well as it different modifications such as adaptive PCM, pulse wide modulation (PWM), IrDA standard for a high-speed communication. In the platform development we choose the PCM-based approach for remote control and inter-robot communication that provides a half-duplex data exchange.

\textbf{2.} The level of communication protocols concerns the propagation of information in a swarm. The main problem represents a routing of information packages, which requires such computational capabilities that a small microrobot mostly does not possess.

\textbf{3.} The level of information structures includes issues such as a software architecture, a choice subsystems that require a robot-robot communication,  optimization of communication flow, an optimal representation of information, availability of information and others.

\textbf{4.} The last level concerns a collective behavior of the whole swarm. This issue is specific for mobile collective systems with a limited communication range. For a large-distance (comparable with a size of the whole system) communication, robots need multiple peer-to-peer connections. This dynamical network can exist in such scenarios when the robots operate in small clusters (\emph{small area swarms}), when robots move between clusters (\emph{large area swarms with inter-clusters exchange}) or robots purposefully create and maintain the communication network.

Usually, messages that require propagation trough a swarm have a spatial context: resources, events in a specific position, calculation of spatial values (size, length) and others. In these cases the content of information (i.e. about the event itself) is not relevant, more important is its spatial context (i.e. where it happens). Only a small number of possible events is non-spatial: e.g. messages used for a global coordination (such as "job is done") or collective decision making~\cite{Kornienko_OS01}. The package-based approach, see e.g. \cite{Zhang1098}, does not provide any accounts about spatial origin of an event, since the information is propagated through a sequence of robots. In order to achieve a sender, a robot has to follow the propagation way, which quickly disappears because of movement of robots. In two following sections we consider virtual fields and epidemic algorithms, which are useful not only for communicating a content but also for providing a context of events.

\subsection{Virtual Fields}

Virtual field is a general name for different approaches, which use only a single value for communication, so-called attribute (such as intensity, connectivity, degree of clusterization) and spread information about this attribute over the whole system. Since this information is updated locally, there appear local inhomogeneities, which provide a context of information. Typical example is a pheromone-based approach~\cite{Payton01}, which is referred in the context of emergent phenomena~\cite{Kornienko_S04b}, \cite{Kornienko_S04a}, artificial evolution~\cite{Kernbach08online} or robotics~\cite{Kernbach08Permis}.

Pheromone-based communication originates from the insect world. There are many works devoted to this phenomenon in natural systems as well as to applications in technical systems. Pheromone-based communication can be divided into two main groups: pheromone leaved on immovable objects (ground, floor) and pheromone leaved on moving objects such as robots. The first type of pheromone assumes usually real (physical) pheromone, such as chemical substances or electromagnetic marks, the second type of pheromone can also have a virtual nature. For instance, robots exchange the values of variables; since these variables are "located" in a robot we speak about "virtual pheromones".

Propagation of pheromones creates a spatial field, which can be of four different types: \emph{non-gradient} (used for transmitting some signals), \emph{gradient} (to provide spatial context of a message), \emph{directional} (propagated in a specific direction), \emph{functional} (e.g. repelling or attracting). The pheromone itself can be calculated as a function of connectivity (the number of neighbors, see~\cite{Nembrini02}), time, specific input (e.g. only robots with a specific sensor input can transmit a pheromone) or embodied information. Pheromone field can consist of one or many different pheromones or even of different subfields, i.e. with hierarchical structure and be used for different complex actives such as an adaptive control of cooperative locomotion~\cite{kernbach09adaptive}.

Pheromone field provides spatial information about different events in a swarm. This spatial context can be useful not only for information transfer, but also for many other activities such as navigation, localization or even a spatial or temporal planing~\cite{Kornienko_S03A}. For instance, robots can perform triangulation with three pheromone sources, in a similar manner with GPS. The sources of a pheromone can be also such robots that encounter a relevant event for a swarm and serve as orientation points. Generally, the type of information, circulated in a swarm, depends on the type of activity: in a spatial type -- information about orientation and position of robots/objects, information-based -- e.g. sensor data or internal states~\cite{Kornienko_S05b}. In Table~\ref{tab_context} we collect some examples of context and retrieving mechanisms.
\begin{table}[ht]
\centering
\begin{tabular}{l | l | l  } \hline
{\bf Context}    & \textbf{Can be useful for} &{\bf
Requirement}\\\hline
direction of     &   - building gradient mechanisms  &  directional  \\
received message &   - directional diffusion fields  & communication\\
                 &   - spatial awareness           \\ \hline
signal intensity &   - distance to communic. neighbor  & analyzing  \\
                 &   - building gradient mechanisms  & of ADC signal\\\hline
noise during     &   connectivity, many noise $\rightarrow$  & noise filter\\
communication    &   many robots around\\\hline
difference in    &  ambient light  & analyzing \\
logic "0/1" signal &  (specific for IR-communication) & of ADC signal\\\hline
N of communic.   & communication density & storing \\
requests         &  & of requests\\\hline
hop counter, ~\cite{Geider06}& the swarm size and  & sending hop\\

                 & swarm density & counter\\\hline
\end{tabular}
\caption{\small Some examples of a message context.} \label{tab_context}
\end{table}

To give an example, let us assume, a robot has found a "source" being relevant for the whole swarm. It sends the message "I have food source" coded numerically. When this message is propagated through a swarm, each robot knows "a resource is found by some robot". However, robots cannot find it because they do not know a spatial position of this "food source". The initial robot cannot provide these coordinates because it does not know its own position. However robots, when receiving this message, can estimate the distance and direction to a sender and the number of neighbors, which also received this packages. This information is not contained in the message itself, robots have to retrieve this context information from different sources.
Now by using this context information robots can apply different algorithms to find the origin of messages, e.g. to follow the gradient as suggested in Sec.~\ref{sec:virtualFieldsEx}.

\subsection{Epidemic Algorithms}

Information in the case of epidemic algorithms is spread in a swarm in the infection-like manner. The main difference to the pheromone-based approach is that each message is propagated further without any updates with essentially lower re-sending rate. Thus, it does not create gradient fields. For further references to epidemic algorithms we refer to the overview~\cite{ma2009}, spatial and mobility issues are considered e.g. in~\cite{Lloyd1996}, \cite{Chaintreau:2009}. Recently, a number of publications considers epidemic approaches in the context of evolutionary robotics~\cite{Kernbach09Platform}.

\begin{figure}[t]
\centering
\includegraphics[width=.8\textwidth]{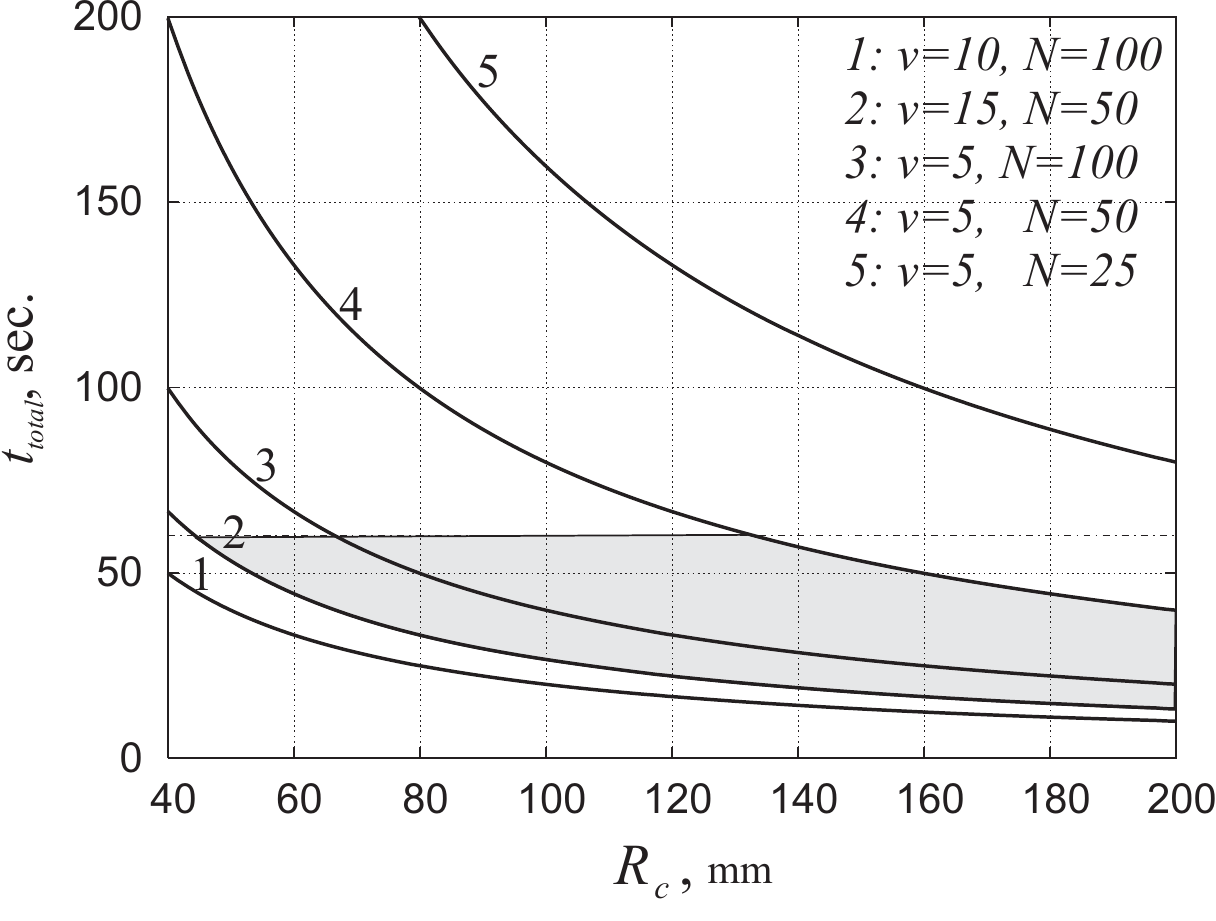}
\caption{\small Total propagation time $t_{total}$ as a function of communication distance $R_c$ with different values of velocity $v$ and the number of robots $N$. Area available for the swarm is $1000\times1000~mm^2$.
\label{fig:communication_radius}}
\end{figure}

Since epidemic algorithms require typically a low number of messages, the question is how quickly these messages are spread in a swarm. This can be estimated when we know how many communication contacts $n_c$ will happen during the motion of a robot. This value is equal to the average number of robots in the area $S_c$ and can be expressed as~\cite{Kornienko_S06}:
\begin{equation}
n_c=\frac{2 \sqrt{2} R_c \upsilon t N}{S_{sw}},
\label{eq_nc2}
\end{equation}
where $R_c$ is the communication radius, $\upsilon$ -- velocity of motion, $S_{sw}$ swarm area, $N$ -- number of robots. The dynamics can be easily estimated in the telescopic way:
\begin{enumerate}
    \item In the first step the first robot "infects" $n_c$ robots
    (the first generation). The number of "infected" robots in this step is $n_c+1$

    \item In the second step the first robot "infects" again $n_c$ robots
    (the second generation). The $n_c$ "infected" robots from the first step
    infect $n_c n_c$. The number of "infected" robots in this step: $n_c + n_c^2$

    \item In the third step the first robot "infects" $n_c$ robots. The
    the second generation of robots "infects" $n_c n_c$. The the first
    generation of robots "infect" again $n_c n_c$ and the "infected"
    by them robots "infect" in turn already $n_c n_c n_c$. Number of "infected"
    robots in this step $n_c+n_c^2+(n_c^2+n_c^3)$.
\end{enumerate}
Collecting this telescopic dynamics, we obtain
\begin{eqnarray}
&&[n_c+1] + [n_c+n_c^2]+ [n_c+n_c^2+(n_c^2+n_c^3)]+...= \\
&&[n_c+1] + n_c[n_c+1]+ n_c[n_c+1+n_c(n_c+1)] +...
\end{eqnarray}
This relation can be written iteratively as
\begin{equation}
k_n= k_{n-1}+n_c k_{n-1}=k_{n-1}(n_c+1), ~~~~~ k_0=1,
\end{equation}
i.e. it possess an exponential form $(n_c+1)^n$. Now we are interested in the case when all robots are "infected" $(n_c+1)^n \geq N$ or $n=log_{(n_c+1)}N$. The information transfer start if the first robot "infects" one additional robot; the time until the first infection can be obtained as
\begin{equation}
t_{first}= \frac{S_{sw}}{2 \sqrt{2} R_c \upsilon N},
\end{equation}
whereas the total time $t_{total}= n ~t_{first}$ for infecting the whole swarm is
\begin{equation}
t_{total} = \frac{S_{sw}}{2 \sqrt{2} R_c \upsilon N}~log_2(N). 
\label{t_total}
\end{equation}
Several cases of this dependency is shown in Fig.~\ref{fig:communication_radius}, where we observe a fast distribution of information over the swarm.

Despite both the virtual fields and epidemic algorithms spread information in the system, their application is different. Virtual fields require a high-rate sending of updates, as shown in Sec.~\ref{sec:virtualFieldsEx}; they are computationally intensive not only for individual robots but also for the whole swarm. Epidemic algorithms require sending messages with very slow update rate or even only one time, see Sec.~\ref{sec:epidemicExample}, and calculate population or temporal aspects of the macroscopic dynamics locally. Specialization on spatial and temporal phenomena is typical to some extent for virtual fields and epidemic algorithms. Before we provide examples for both approaches, we need to consider embodiment issues such as communication and sensing radii, number of communication channels, covering rate and others.

\section{Hardware Implementation}
\label{sec:hardware}

To implement the mentioned approaches we need to consider the embodiment in terms of a relationship between macroscopic features and hardware constraints~\cite{Kernbach08_2}, \cite{Suzuki95}, and generally in terms of hardware design~\cite{Levi10}. As proposed in~\cite{Kornienko_S05d}, we use the IR signal transmission system. Since the used sensors have the most intensive impact on the features of the communication, in Table~\ref{tab pur_sens} we collect the tested IR sensors with some parameters from their datasheets.
\begin{table}[h!]
\centering
\begin{tabular}{p{4.5cm}@{\extracolsep{3mm}} | c | c | c} \hline
{\bf IR device }              & {\bf wavelength,} & {\bf opening } & {\bf reflection} \\
                              & nm               &  \textbf{angle}& \textbf{dist.},mm \\
\hline \hline
IR receivers (photo-transistor) \\[3mm]\hline
LPT80A   & 470-1080   & 70 & ---\\
TEST2600 & 870...1050 & 60/120 & ---\\
TEFT4300 & 875...1000 & 60 & ---\\
SFH310   & 380...1080 & 50 & ---\\
SFH3100F & 840...1080 & 30 & ---\\[2mm]
TSOP1836 (TSOP4836) & 850...1050 & 90 & ---\\
         & PCM 36 kHz&    & \\
[3mm]\hline
IR emitter (rad. intens. $I_F=100$mA) \\[3mm]\hline
TSSS2600 (1.5 mW/sr)    & 950 & 50/120 & ---\\
IRL80A (1 mW/sr)        & 950 & 60 & --- \\
TSKS5400-FSZ (2-7 mW/sr)& 950 & 60 & --- \\
LD271L  (15 mW/sr)      & 950 & 50 & --- \\
IRL81A (5 mW/sr)        & 880 & 50 & --- \\
SFH409 (6.3-12 mW/sr)   & 950 & 40 & --- \\
TSHA 6203 (25-40 mW/sr) & 870 & 24 & --- \\
SFH4510 (50 mW/sr)      & 950 & 28 & --- \\
LD274  (50 mW/sr)       & 950 & 20 & --- \\
TSAL6100 (80 mW/sr)     & 950 & 20 & --- \\
SFH484 (50 mW/sr)       & 880 & 16 & --- \\
[3mm]\hline
Reflective sensors \\[3mm]\hline
SFH9201        & 900/950 & --- & 1-5 \\
TCNT1000       & 950     & 45 & 0.5-5\\
TCRT 1000/1010 & 950     & 45 & 1-10\\
GP2D120        &         & & 40-300 \\
QRB1134        &         & & 10-500 \\
QRB1113        &         & & 10-500 \\\hline
\end{tabular}
\caption{\small Different IR sensors used in experiments with virtual pheromones and infection dynamics.
\label{tab pur_sens}}
\end{table}

\begin{figure}[ht]
\centering
\subfigure[\label{fig_model_cube}]{\includegraphics[width=.49\textwidth]{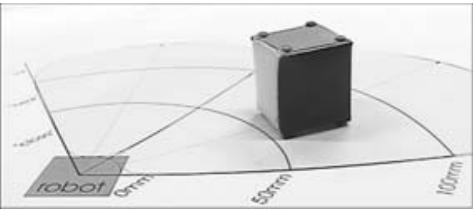}}~
\subfigure[\label{fig_proximity_50_100}]{\includegraphics[width=.49\textwidth]{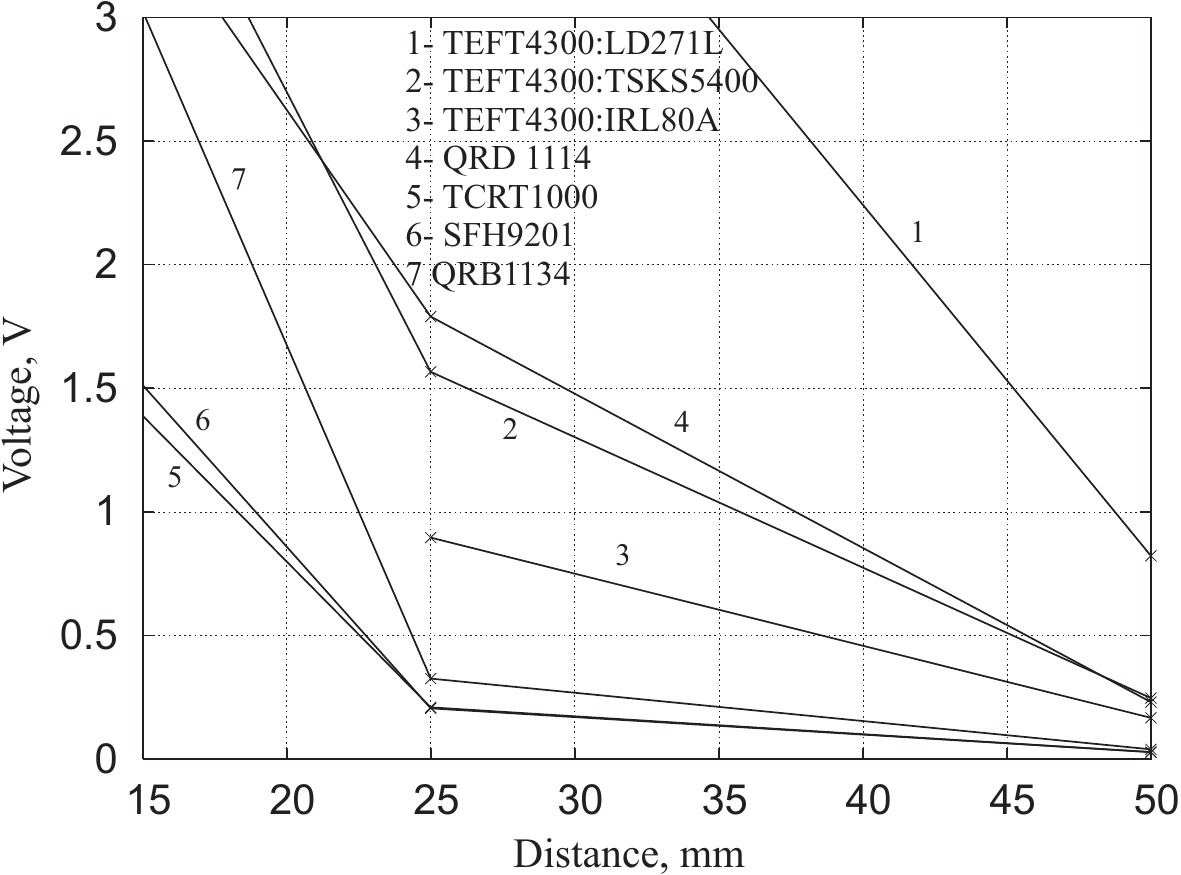}}
\caption{\small \textbf{(a)} Plastic model of a micro-robot with different reflective surfaces used in experiments. {\bf (b)} Degradation of
$V_o$ at shifting an object from the central line in the distance of 100 mm.}
\end{figure}

In the signal transmission the following parameters of IR sensors are important: max. measuring distance $R_{max}$, optimal recognition distance $R_{rec}$, opening angle of radiation/reflection ray $\alpha$ on $R_{rec}$, degradation of radiation outsize opening angle $D_{dist}^{in}/D_{dist}^{out}$, dependency of reflection on color/slope of an object, object and geometry resolution $O_{res}$, $G_{res}$ in $R_{rec}$. In experiments we measured them. Obviously, the narrower is the radiation ray, the better is the resolution of perception system. Therefore we choice IR-emitters only with small opening angles, as e.g. LD274, SFH484 and SFH4510. Moreover, we also tested distance sensors, that combines emitter and receiver, such as GP2D120, QRB1134 and QRB1113. For experiments we made a model of a robot as a cube from plastic with the edge 25 mm, see Fig.~\ref{fig_model_cube}.
Sides of this cube are painted in different colors so that we can compare reflectivity depending on objects color. Experiments have been done by measuring a voltage $V_o$ on the emitter of phototransistor (output of distance sensor). The emitter resistance are chosen so that at a maximal reflection the max. voltage equals $V_o\approx5 V$. Measurements have been done with the digital voltmeter "Voltcraft M-3850".

Firstly, we started with distance sensor GP2D120. As stated in its datasheet, it can measure distances between 40 and 300 mm, $R_max=300mm$. The sensor is not sensitive to ambient light, however for open distances (over 300-500 mm), it produces a "background" voltage that depends on illumination. The exact dependency was not established. If the object is placed within 300 mm range, this sensor delivers the values that are independent from a color of the object, slope, and the light. However this sensor, perhaps because of non-symmetrical construction, has different values on left and right part regarding the central line. The distances are measured only in the left part (the emitter part with opening angle 30$^o$), whereas when the object was putted in the right part (receiver part), values of distances are getting "strange", e.g. for the "background" voltage 0.34 V we obtained values 0.04-0.028 V! We cannot explain this behavior, however this makes difficult any further calibration of this sensor.

Next series of experiments has been done with separated IR-emitters and receivers. We choice IR-receivers only with ambient light filter. We are going to use the same receiver for distance measurement and communication, therefore we prefer sensors wide opening angle, e.g. TEFT4300, TEST2600 ($\alpha=60$), in the "control group" we have SFH3100F with $\alpha=30$. The current $I_F$ of IR-emitter is restricted to 20 mA ($V_{cc}=5V$) that corresponds to max current of I/O port. We draw in each "zone" one curve (50mm, 100mm, 150mm), and put on this curve points with the interval of 12,5mm. Model was shifted each time to the next point (this corresponds to half-body displacement), where we performed the next series of measurement for white and black side of model. For each pair of IR-emitter-receiver we get $\approx$ 50 values. The minimal recognizable difference between two voltage values is $\approx$ 0.02 V for 8 bit ADC. In this way, 140-150 mm is the maximal distance $R_max$ of these reflective sensors.

The values in the zone "near" are contained between 0.1 and 1 V, whereas for the zone "close" between 1 and 5 V. The distance of optimal recognition $R_rec$ lies within the zone "near" because here the voltage function is more or less "linear" and object does not "block" sensors in contrast to the zone "close". The minimal distance is about 5 - 10 mm and depends on the structure of the sensor (emitter and receiver are parallel or not). Generally, a detection of touch (contact with an object) is not possible with the reflective IR-sensor (if the sensor is not displaced regarding the contact surface). The voltage $V_o$ in fact does not depend on the slope of cube object, however sensitive to the color. For the black side of the model, the $V_o$ was reduced by a factor of five or even more. Other colors, though they also change $V_o$, do not have such a deep impact. Therefore, the robot has to be of white or, at least, light (grey) color.

The sensor cannot differentiate whether the object is putted in the central line but on a large distance, or the real distance is smaller but the object is displaced from the central line. To solve this problem, we suggest two approaches. Firstly, we can preform two measurements, with proximity and distance sensors. Wide opened proximity sensor provides an approximative distance, whereas distance sensor, based on this, can estimate a displacement. In the second approach, robot can turn so that to obtain maximal $V_o$ (object on the central line). Obviously, this approach requires more time, and therefore is more suitable for immovable objects.

Since a direct receiver-emitter optical connection provides essentially more IR-radiation than a reflective optical connection, communication is less problematical than a distance measurement for IR devices. In this way, if proximity sensors are able to send and receive a reflective signal within the "near" zone, they can send and receive a direct signal within the "far" zone. However, we need to provide 60$^o$ opening angle for a multi-channel directional communication system.

In experiments we used the following pairs TEST2600:TSSS2600, TEFT4300: (IRL80A, TSKS5400-FSZ, LD271L), integrated sensors SFH9201, TCNT1000, TCRT1000, QRB1134, QRD1113. Generally we tested also IR-emitters with small opening angles like (SFH409), but they do not satisfy the requirements. We also have several problems to isolate TEST2600:TSSS2600 optically one from another. This pair has wide vertical opening angle 120$^o$, so that to remove completely a leak of IR-radiation in sensor was not really possible and we caused after many tries experiments with this pair. In Fig.~\ref{fig_proximity_50_100} we demonstrate the emitter voltage of IR-receiver in dependence of distances in "near" and "close" zones. We have here three different groups. The first group composes IR-emitters with more or less narrow opening angle. Due to a smaller angle, they have a stronger radiant intensity (example LD271L in Fig.~\ref{fig_proximity_50_100}). The second group is built by the "normal" 60$^o$ emitters IRL80A, TSKS5400-FSZ and the integrated sensor QRD1114. Finally, the integrated sensors SFH9201, TCRT1000, QRB1134 (TCNT1000 demonstrated very small measuring distances and was not considered at all) are included into the last group.

Analyzing the performed experiments, we came to the conclusion that the third group (integrated sensors) is not really suitable for this application, although they have good coverage in 60$^o$ sector. The measured distances start only from 40-50 mm (on the brink of recognizability), and communication radius $R_c$ is about 60-70 mm (also on the brink of recognizability). The first group is also not suitable, because of a poor coverage in 60$^o$ sector, especially of IR-emitters with 40 and less degree. Therefore a compromise represents the second group. From the tested IR-emitters only one TSKS5400-FSZ demonstrated acceptable coverage that can be approximated in the algorithmic way. The sensor QRD1113 shows really good results, however it was extremely sensitive even to luminescent light, so that its further calibration represents essential difficulties.

Tests of communication was performed by means "send impulse"-"receive impulse". Generally, communication results of all pairs from the second group are good. The signal from 150 mm distance on the direct line was of 0.7-0.8 V, in different directions with 60$^o$ of receiver and emitter not less than 0.1 V. Very important is that receiver and emitter are optically isolated so that to provide only 60$^o$ opening angle (they can perceive and send till 80-90 grad). The signal outside of 60$^o$ was less than 0.1 V for sensors with optical isolation. In this way robots can receive very exact information about a spatial origin of signal. Communication distance can be easy reduced (or even increased) in the algorithmic way by putting some threshold on the ADC values of a sensor.

\begin{figure}[t]
\centering
\subfigure[]{\includegraphics[width=.29\textwidth]{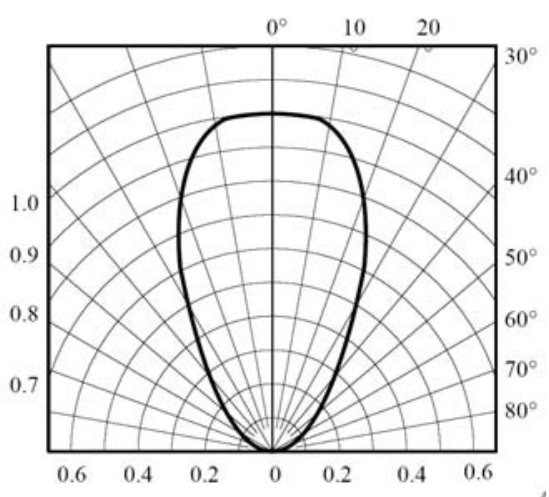}}
\subfigure[]{\includegraphics[width=.29\textwidth]{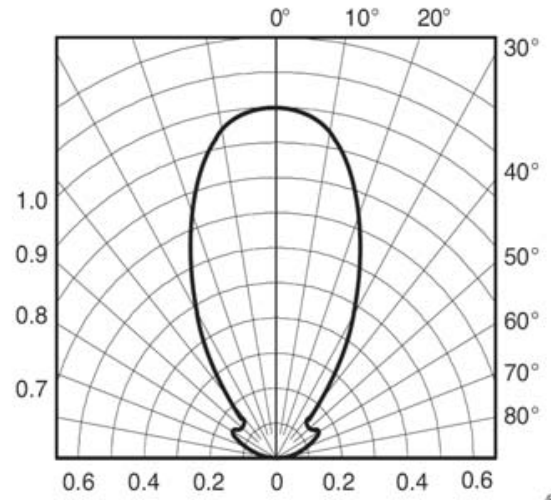}}
\subfigure[]{\includegraphics[width=.38\textwidth]{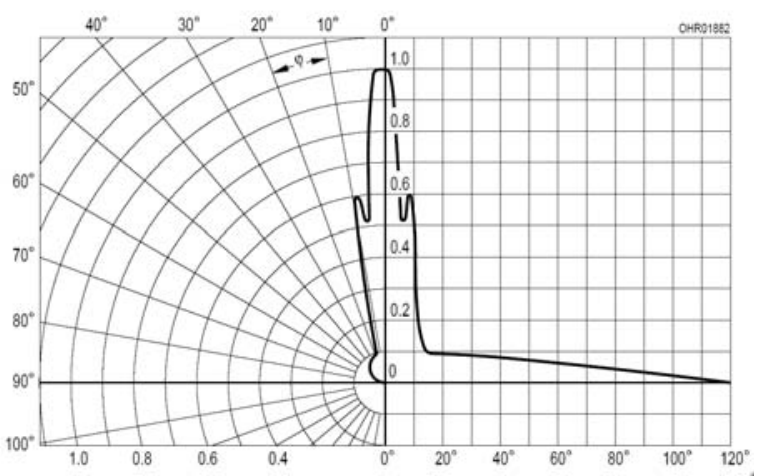}}
\caption{\small
{\bf (a)} Relative sensitivity of phototransistor TEFT4300;
{\bf (b)} Relative radian intensity of IR-diode TSKS5400;
{\bf (c)} Relative radian intensity of IR-diode LD274;
\label{fig_opening_angle}}
\end{figure}

To conclude, the integrated transistor-diode and distance sensors are not suitable, primarily because of the size and the consumed current. In the tested phototransistors with 60$^o$ angle, we can choose TEFT4300 with the collector light current 3,2 mA as the most suitable IR-receiver both for distance and proximity sensing. From this group alternatives are TEKT5400S (72$^o$, 920 nm), PT4800F (70$^o$, 860 nm) SFH 310FA (50$^o$, 880 nm), SFH 300FA (50$^o$, 870 nm) and similar. For the distance sensing suitable are such IR-diodes that are spectrally matched with TEFT4300 and have as small as possible beam angle. In the tested diodes it was LD274 with 20$^o$. Alternatives are TSAL5100\footnote{GaAlAs emitters are generally more suitable than GaAs emitters due to their more higher power output for the same current.} (20$^o$, GaAs/GaAlAs, $>$80 mW/sr), TSTS7100 (10$^o$, 950 nm GaAs, $>$10 mW/sr) and similar. TSKS5400-FSZ demonstrated a good compromise as IR-emitter for proximity measurement and communication. These GaAs diodes emit 950 nm wave length and are suitable also for PCM-coded communication with TSOP1836 (TSOP4836). Alternatives are GL4800 (60$^o$, 950 nm) and similar. In Fig.~\ref{fig_opening_angle} we demonstrate the relative sensitivity and the relative radian intensity of the used
IR-devices.

\begin{figure}[h!]
\centering
\subfigure[]{\includegraphics[width=.49\textwidth]{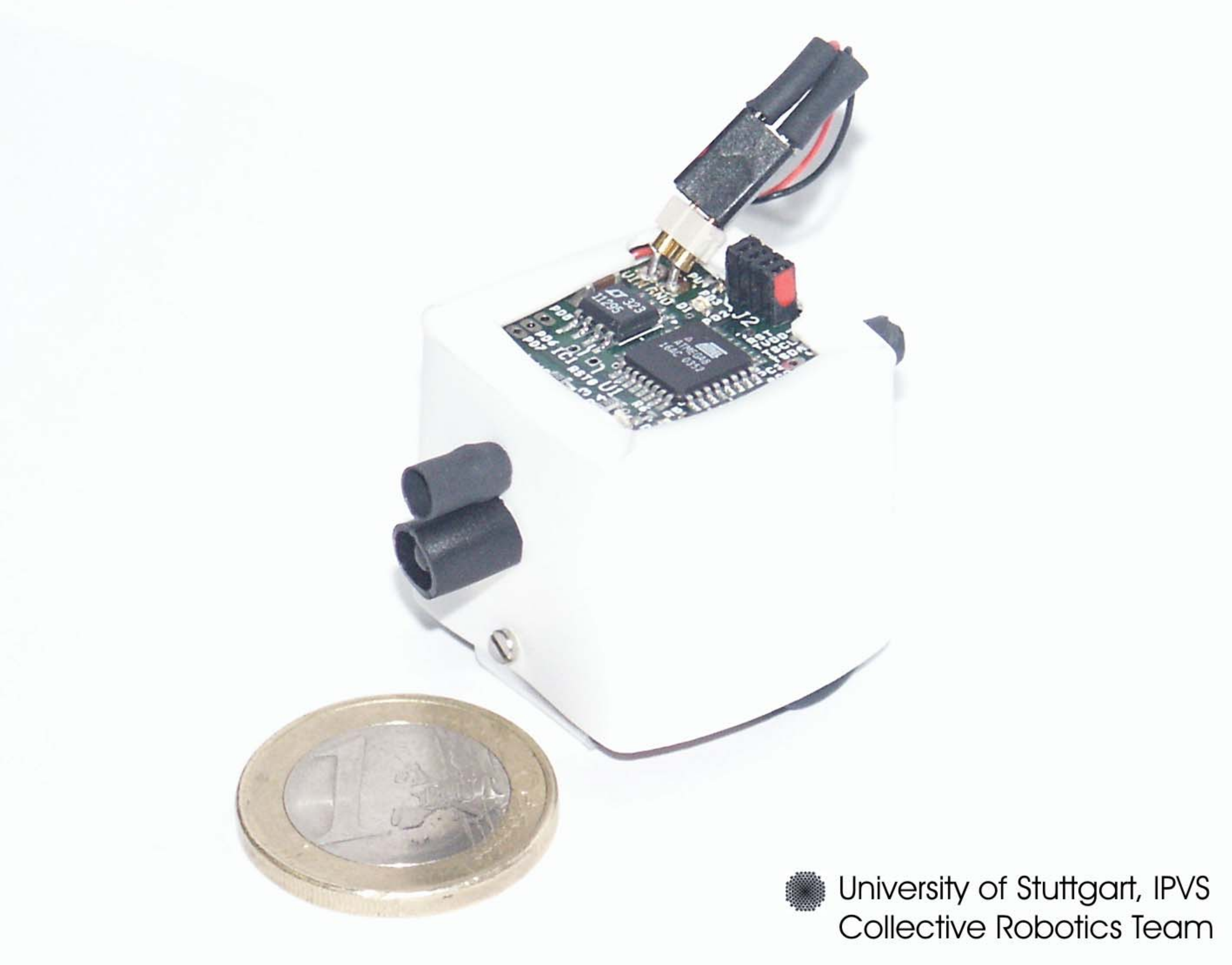}}~
\subfigure[]{\includegraphics[width=.49\textwidth]{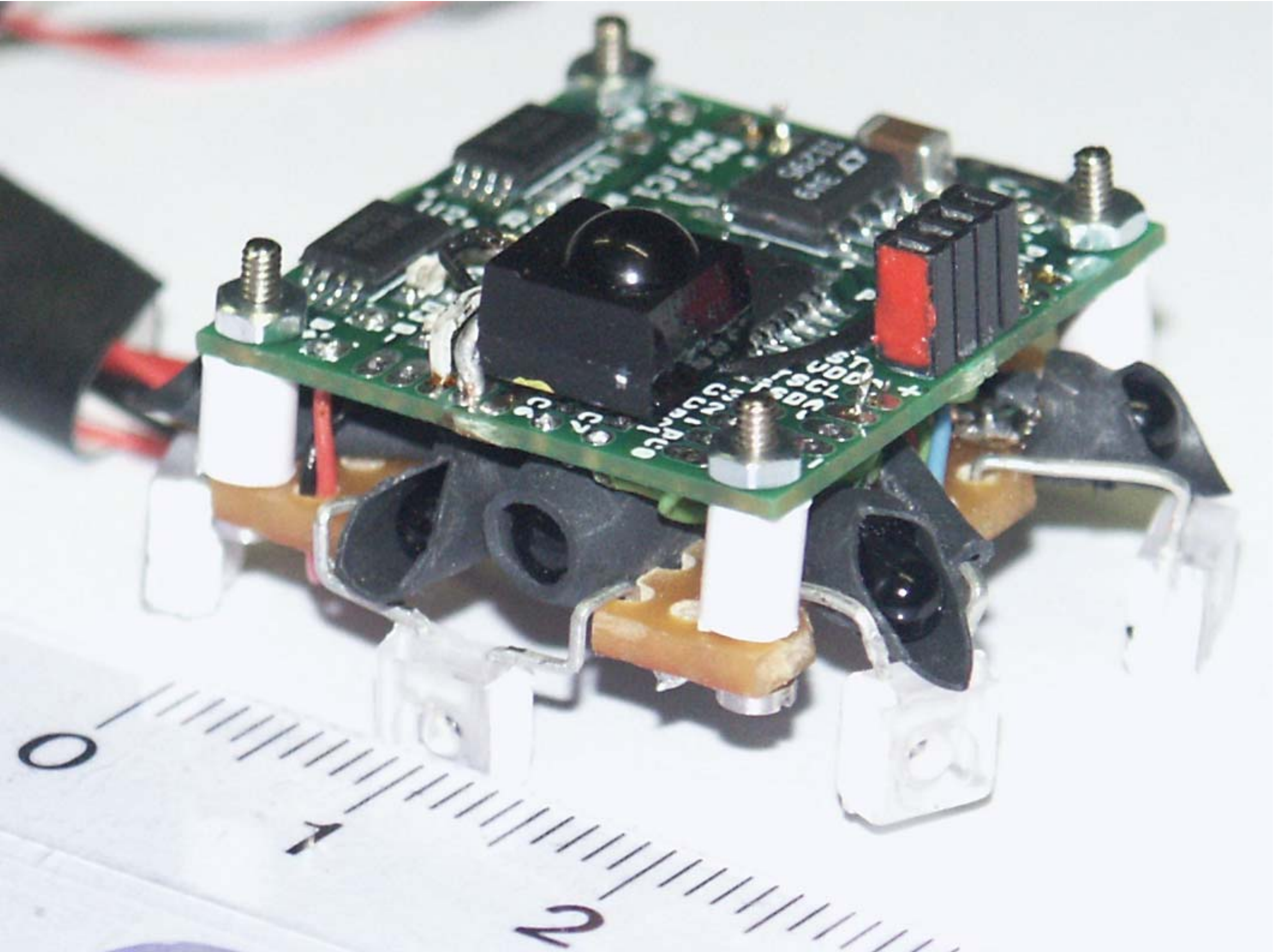}}\\
\subfigure[]{\includegraphics[width=.48\textwidth]{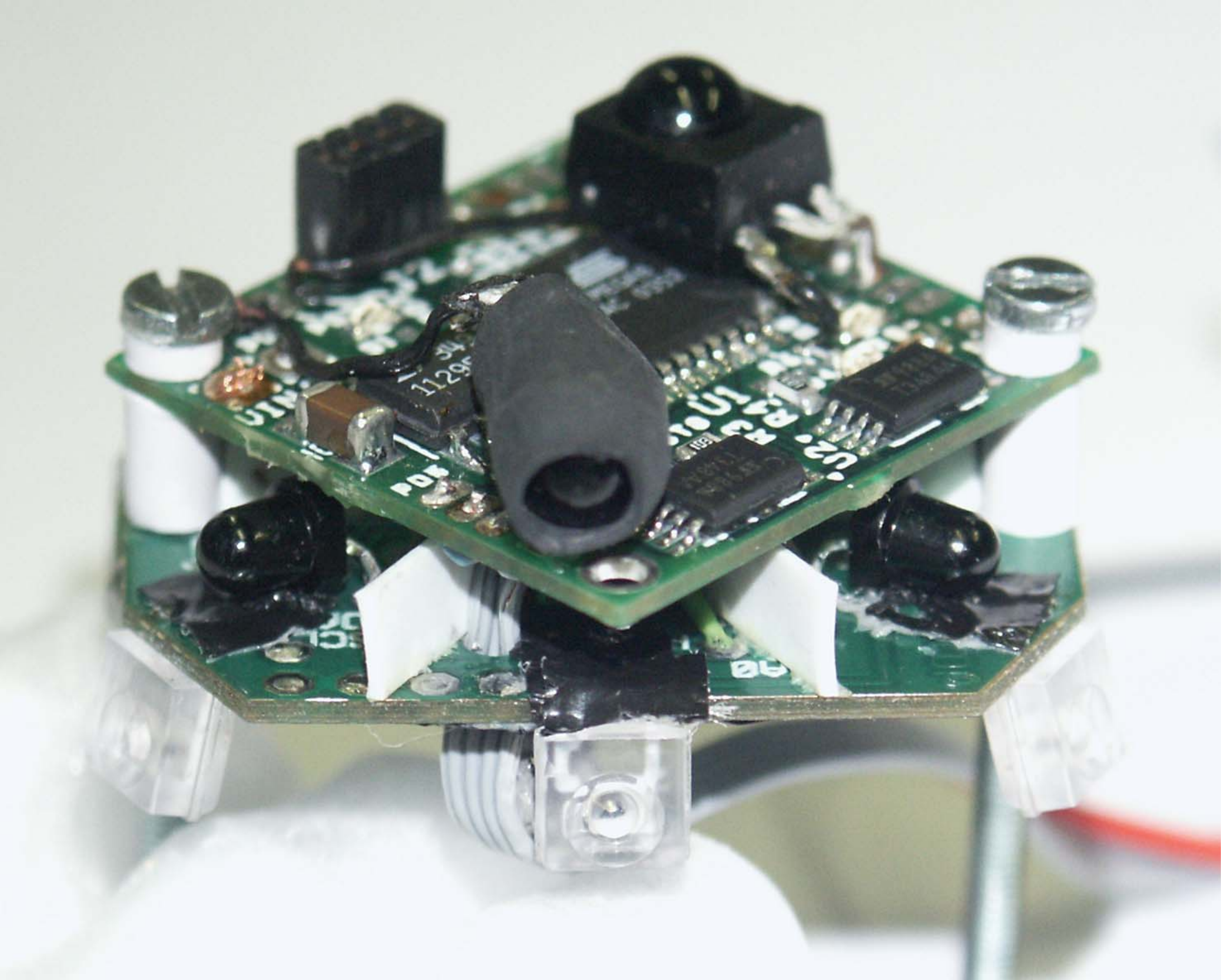}}~
\subfigure[]{\includegraphics[width=.51\textwidth]{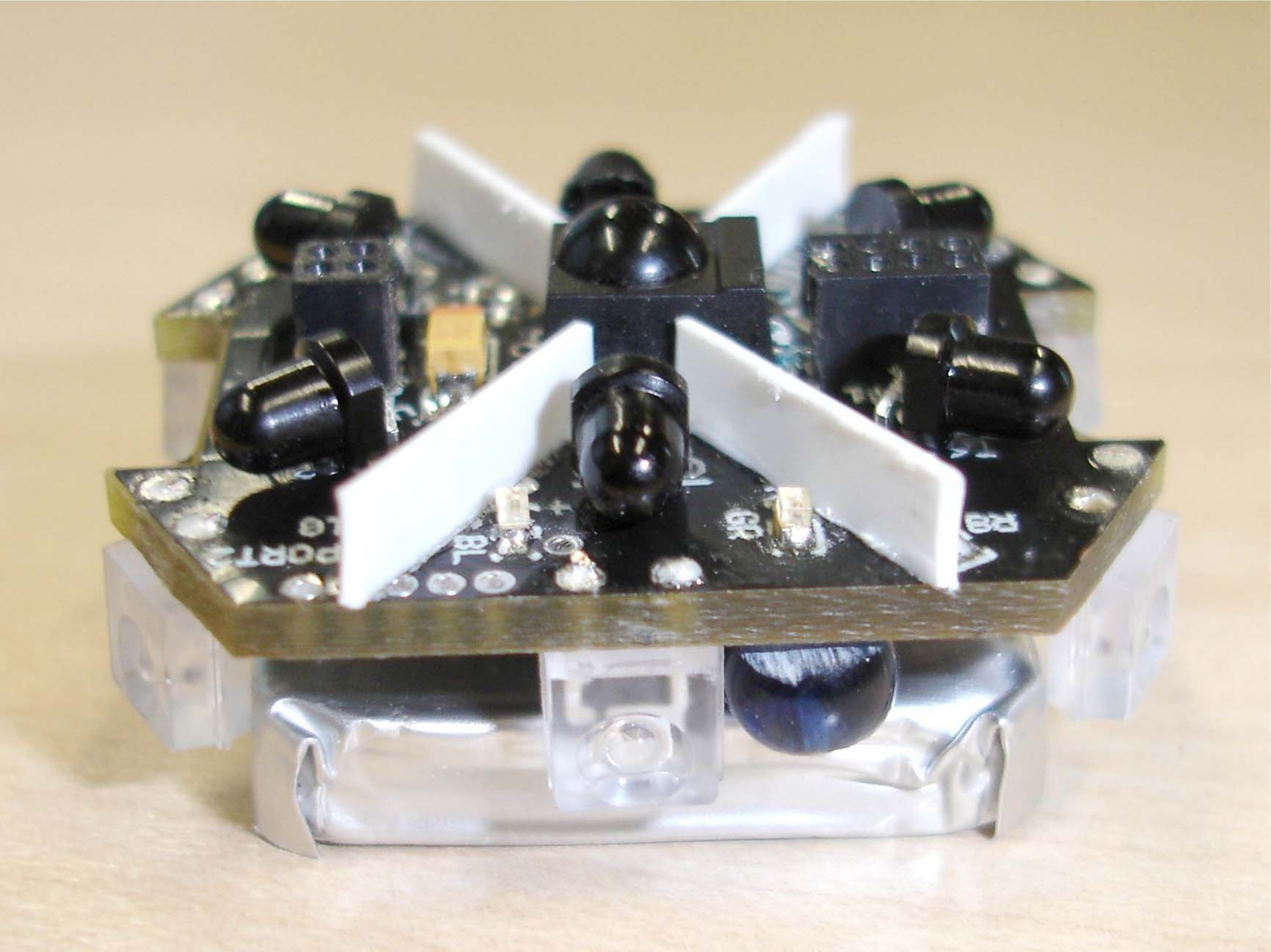}}
\caption{\small
{\bf (a)} The prototype of a micro-robot with sensors and a white covering;
{\bf (b)} The first version of the sensors board (with Megabitty board)
that supports 6-x directional robot-robot and host-robot communication
proximity sensing and perception of surfaces geometry;
{\bf (c)} The second version of the sensors board;
{\bf (c)} The third version of the sensors board.
\label{fig_jasmine-I}}
\end{figure}

The size of IR-devices is important (this factor influences our decision to use TEFT4300 instead of other phototransistors with more higher photocurrent). Unfortunately, we cannot find IR-emitters with 3 mm lens or with side view lens and a radiant intensity of 40-50 mW/sr. Therefore the "distance emitter" (8x$\phi$5 mm) "protrudes" a bit from the chassis of micro-robot, see Fig.~\ref{fig_jasmine-I}.

Proximity sensors are small (emitter 5x5x2.65 mm and receiver 4,5x$\phi$3
mm). As already mentioned, emitters and receiver have to be optically isolated from one another. The most simple solution is to insert a plastic tube on the sensors, as demonstrated in Fig.~\ref{fig_jasmine-I}(a),(b). When the sensors are hidden in the chassis, see Fig.~\ref{fig_jasmine-I}(c),(d), this solution can also provide an optical isolation. The final design shown in Fig.~\ref{fig_jasmine-I}(d) provides a 6x channel directional communication, where sensors are used in the analog mode, i.e. they measure the intensity of IR signals. This feature allows extracting secondary information from a digital signal, which is used in the virtual fields approach.

As a final comment we note that the filament lamps possess also an IR part and thus can be used as a global signal to control a swarm. This does not require any additional sensors, however should used only as an exception, because it essentially distorts a regular communication.

\section{Qualitative Experiments}
\label{sec:Experiment}

Experiments with virtual fields and infection dynamics are performed as a part of other experiments related to navigation, collective perception or decision making in algorithmic~\cite{Kernbach08} and analytic forms~\cite{Levi99} (see \emph{www.swarmrobot.org} for publications related to these experiments). Thus, the experimental data are collected, however they are used in other works. Here, we extrapolate and transform these data to exemplify different mechanisms of information diffusion in a swarm. By this reason, we denote these experiments correspondingly as qualitative experiments.

\subsection{Virtual Fields}
\label{sec:virtualFieldsEx}

The idea of a first experiment is relatively simple: robot X continuously sends a message, say "I found Y". This message is encoded as a number and therefore is compact for transmitting via IR communication system. Any other robot, when getting this message, extract the context, such as intensity of the received signal and direction from where it is received. To introduce a gradient, each robot subtracts one from the numerical value each time when this value (i.e. message) is sent further. This creates a gradient of numerical values, which is maximal at the origin (robot X) and decay towards boundaries of a swarm. Since all robot moves, we introduced a small local amplification: when a message is received within a short distance, i.e. the amplitude of the received IR signal is large, a robot adds one to the message. This creates local inhomogeneities in the intensity of messages and indicates a high swarm density in this location. Each robot stores history of messages; by analyzing this history we expect to discover some global gradient towards a spatial origin of messages and a local gradient explained by the clusterization effect.

\begin{figure}[ht]
\centering
\includegraphics[width=.8\textwidth]{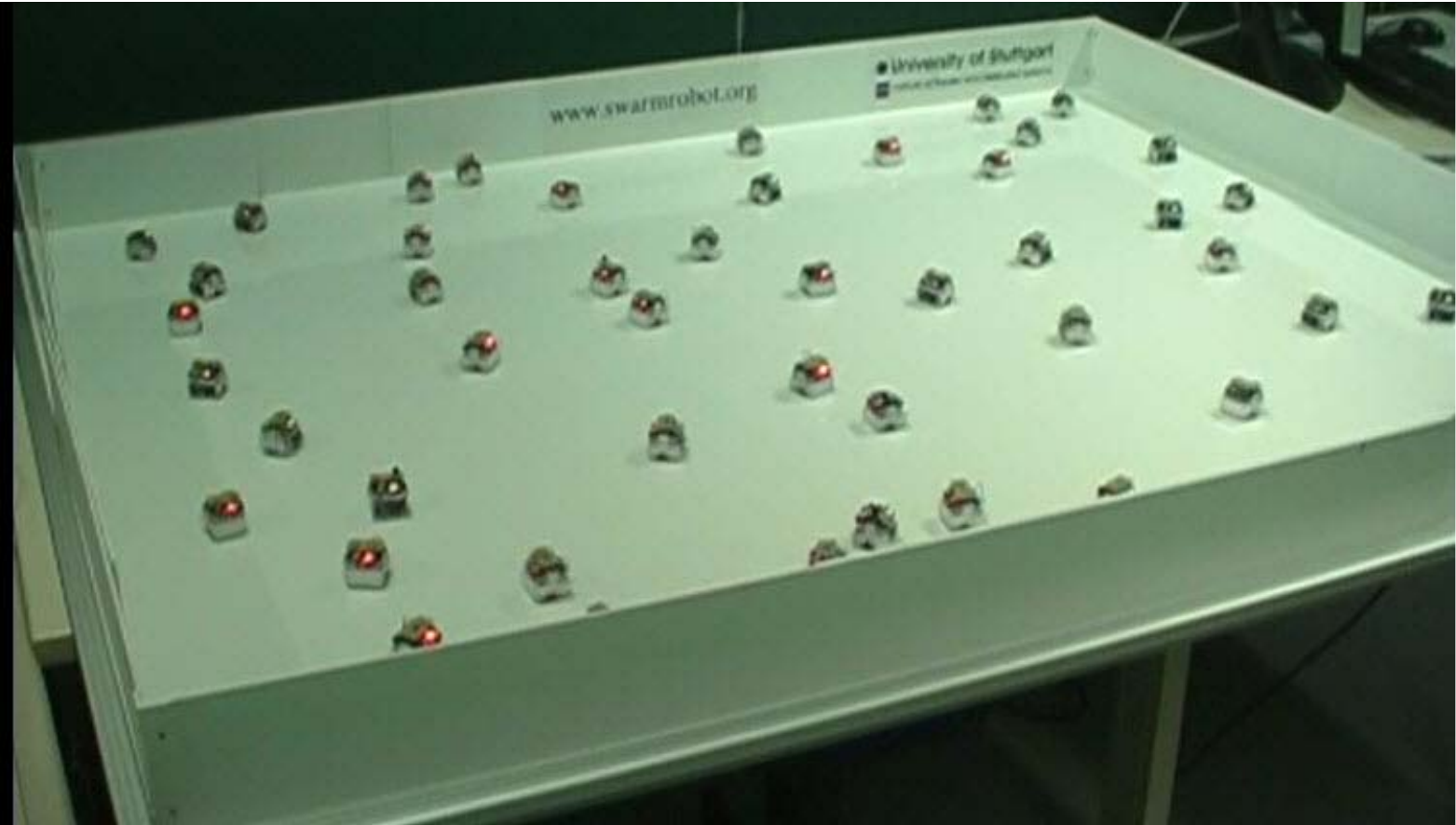}
\caption{\small Setup for the experiment with the virtual field. \label{fig:setup}}
\end{figure}

Since robots are moving, a mapping between spatial positions of a robot at each moment of time and values of the stored messages represents a tough problem. It is resolved by increasing the swarm density higher than a supercritical threshold; in other words robots are catched in a small region by collision avoidance behavior. For the robot arena of $110\times 140cm$ such a number is 50 robots with a large collision detection radius of 15cm, see Fig.~\ref{fig:setup}. Global movement is slow in this case, thus we expect that the original spatial position of a robot will be not essentially changed, when experiment is limited to a few minutes.

\begin{figure}[htp]
\centering
\subfigure[]{\includegraphics[width=.495\textwidth]{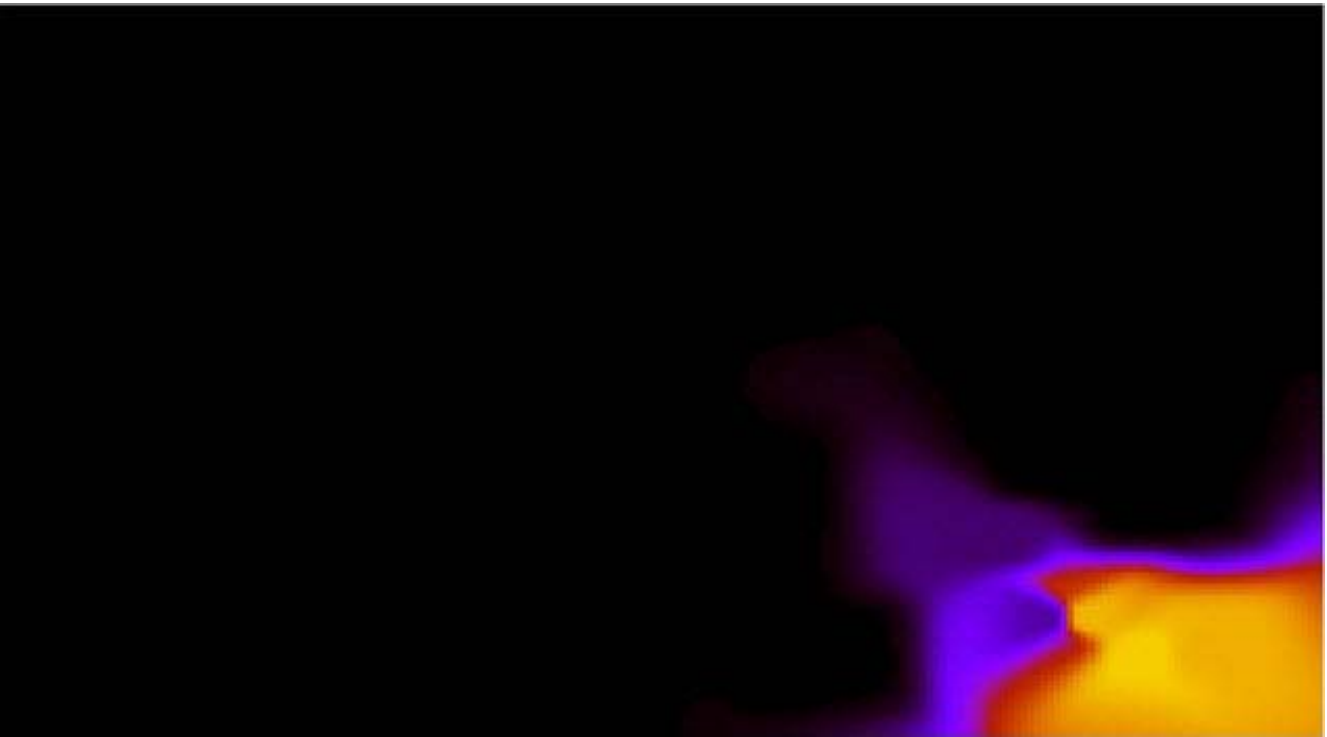}}~
\subfigure[]{\includegraphics[width=.485\textwidth]{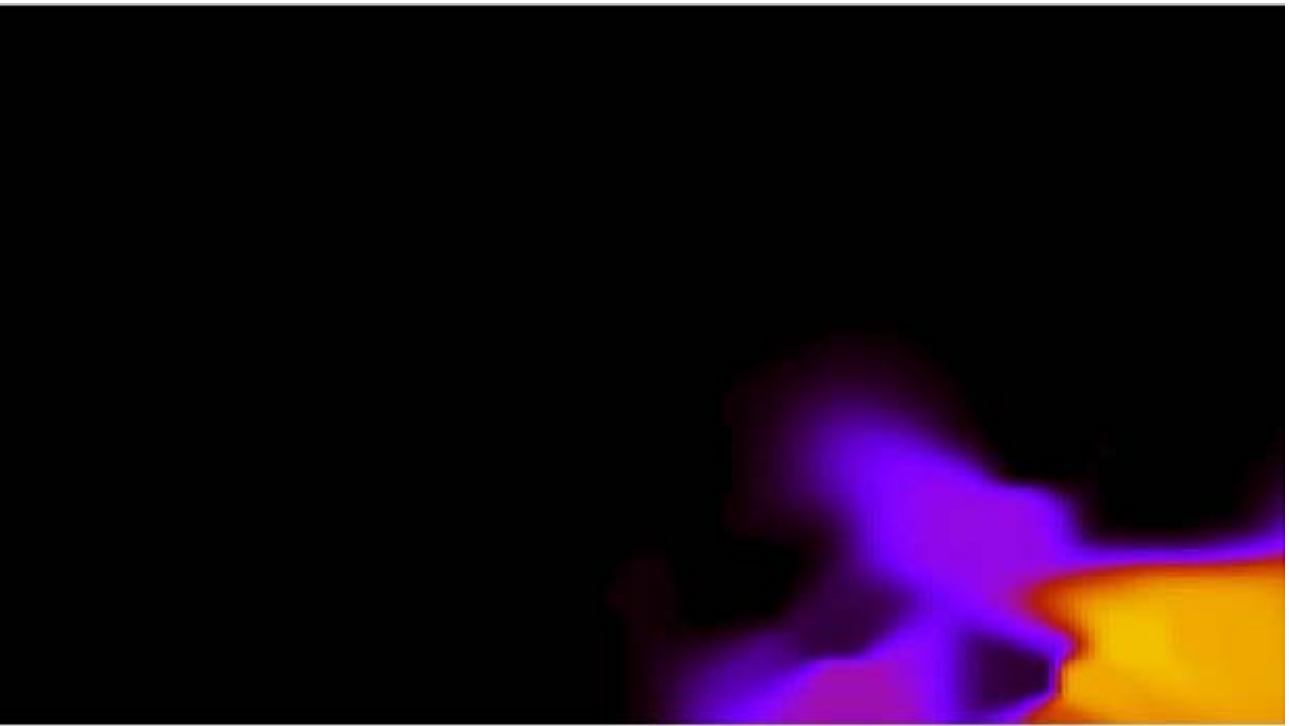}}\\
\subfigure[]{\includegraphics[width=.52\textwidth]{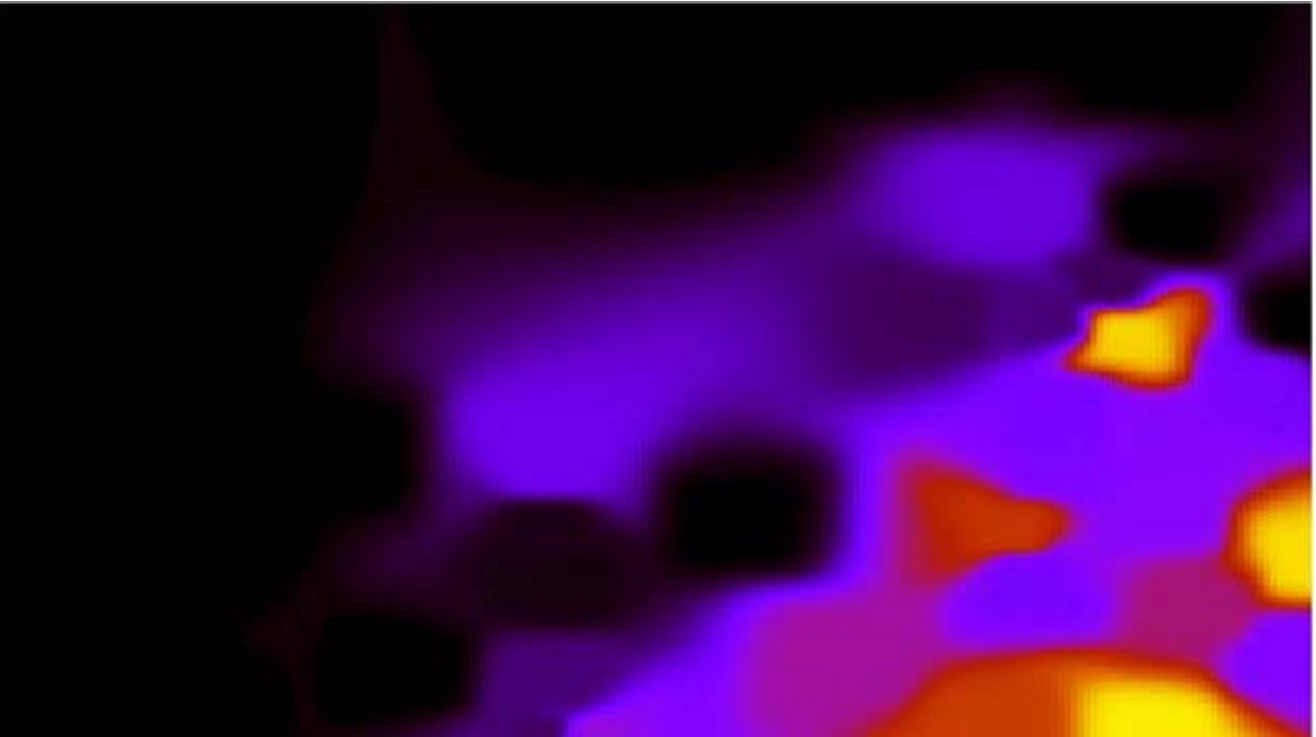}}~
\subfigure[]{\includegraphics[width=.48\textwidth]{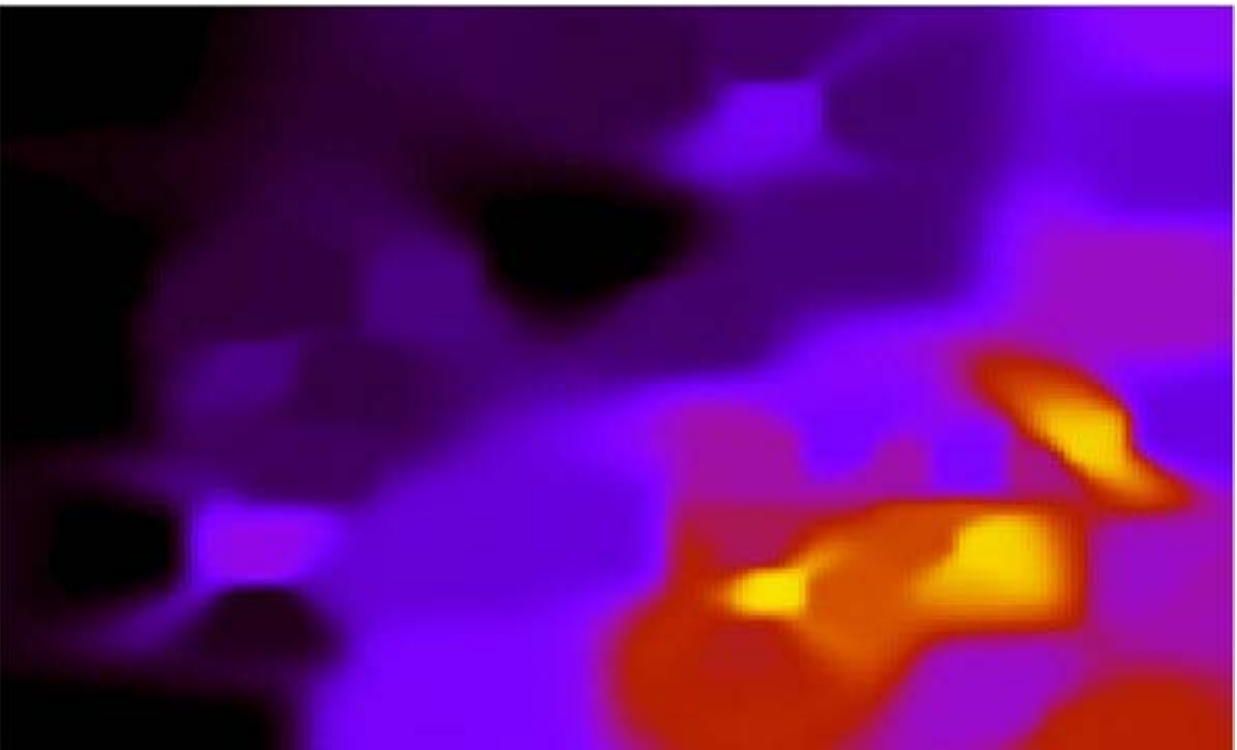}}\\
\subfigure[]{\includegraphics[width=.49\textwidth]{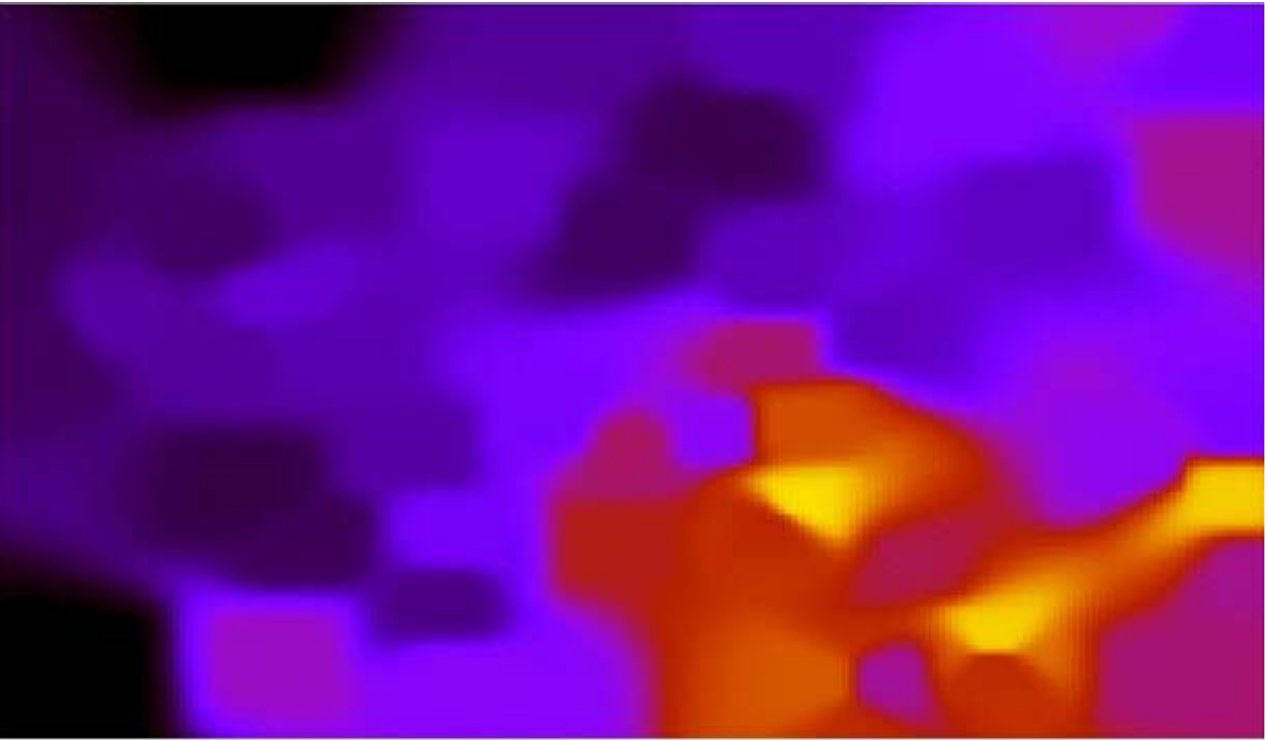}}~
\subfigure[]{\includegraphics[width=.51\textwidth]{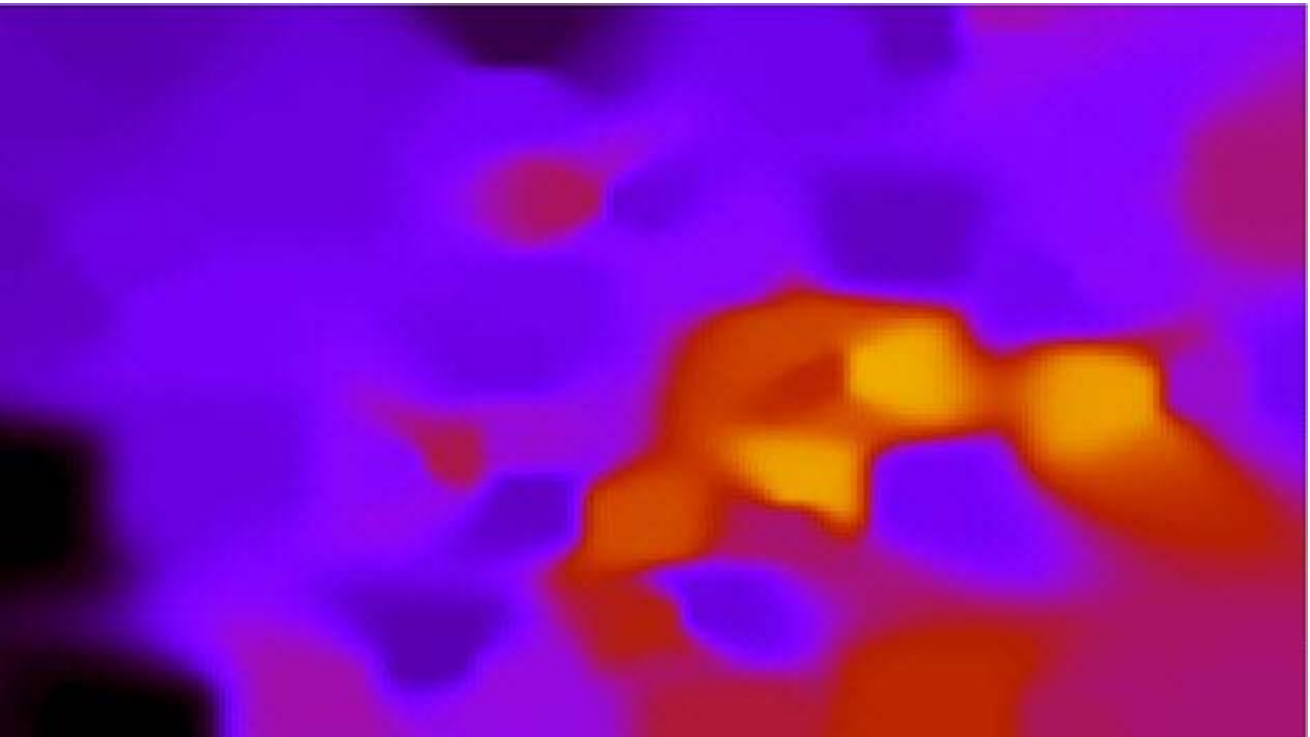}}\\
\subfigure[]{\includegraphics[width=.49\textwidth]{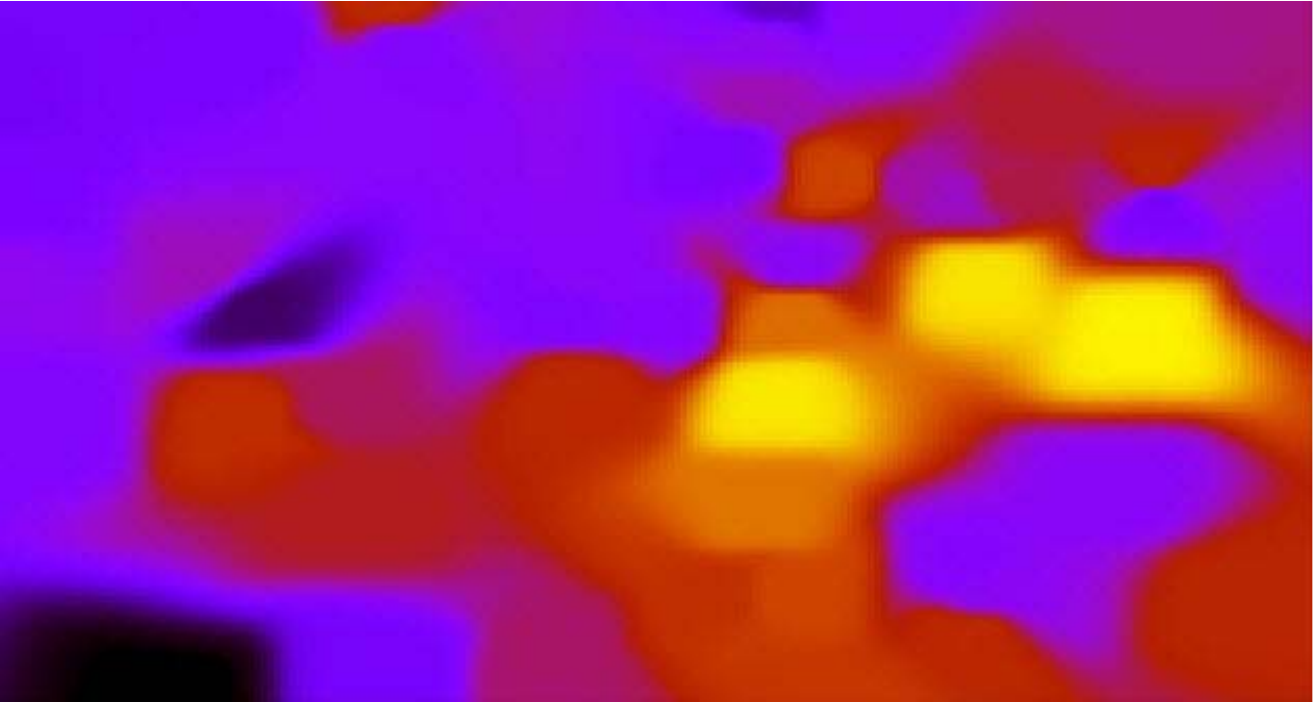}}~
\subfigure[]{\includegraphics[width=.49\textwidth]{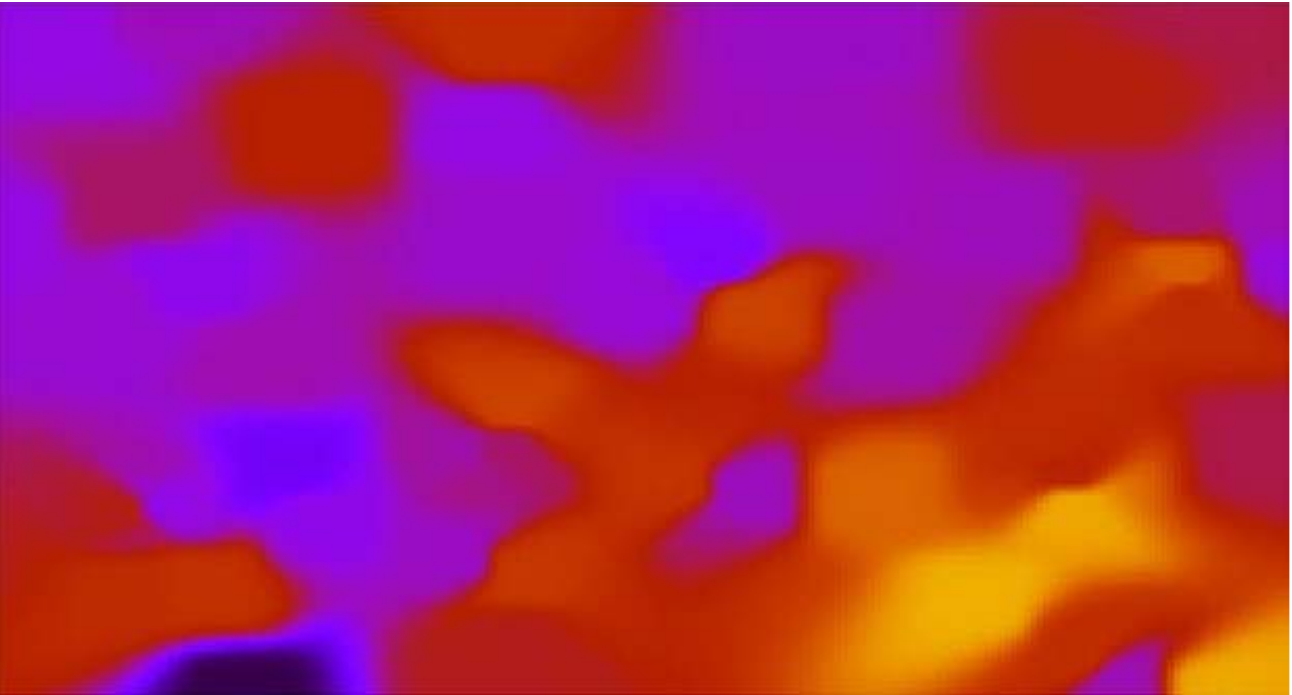}}
\caption{\small Temporal dynamics of the virtual field. Difference between images is 15sec. The robot X (source of the field) is located in the lower right corner.  \label{fig:pheromone2D}}
\end{figure}

This setup was used for decision making experiments, where an initial spatial position of a robot was correlated with a small region on the map. Experiment was performed for two minutes, after the numerical values are read from the memory and displayed as intensities, see Fig.~\ref{fig:pheromone2D}.

The image processing software smoothed edges of the regions. We observe a step-wise propagation of the original message so that all robots become aware of the event. Intensities indicate two effects: first of all we clearly observe a building of temporal clusters, related to a short-range movement of robots. Since the initial "robot X" frequently (approx. 5 times per second) emits the message with a high numerical value and this value decays, we also observe a global gradient between boundaries and the origin.

\subsection{Epidemic Dynamics}
\label{sec:epidemicExample}

The second experiment is performed within a larger series of experiments related to size-dependent aggregation of robots, where we explored the role of modulated and non-modulated signals, IR-noise and other factors. The goal is to introduce a feedback mechanism: the more robots are in the cluster, the longer time such a cluster should survive. It is expected, that small clusters of robots will "travel" around the seed point.

\begin{figure}[ht]
\centering
\subfigure[1   sec.]{\includegraphics[width=.32\textwidth]{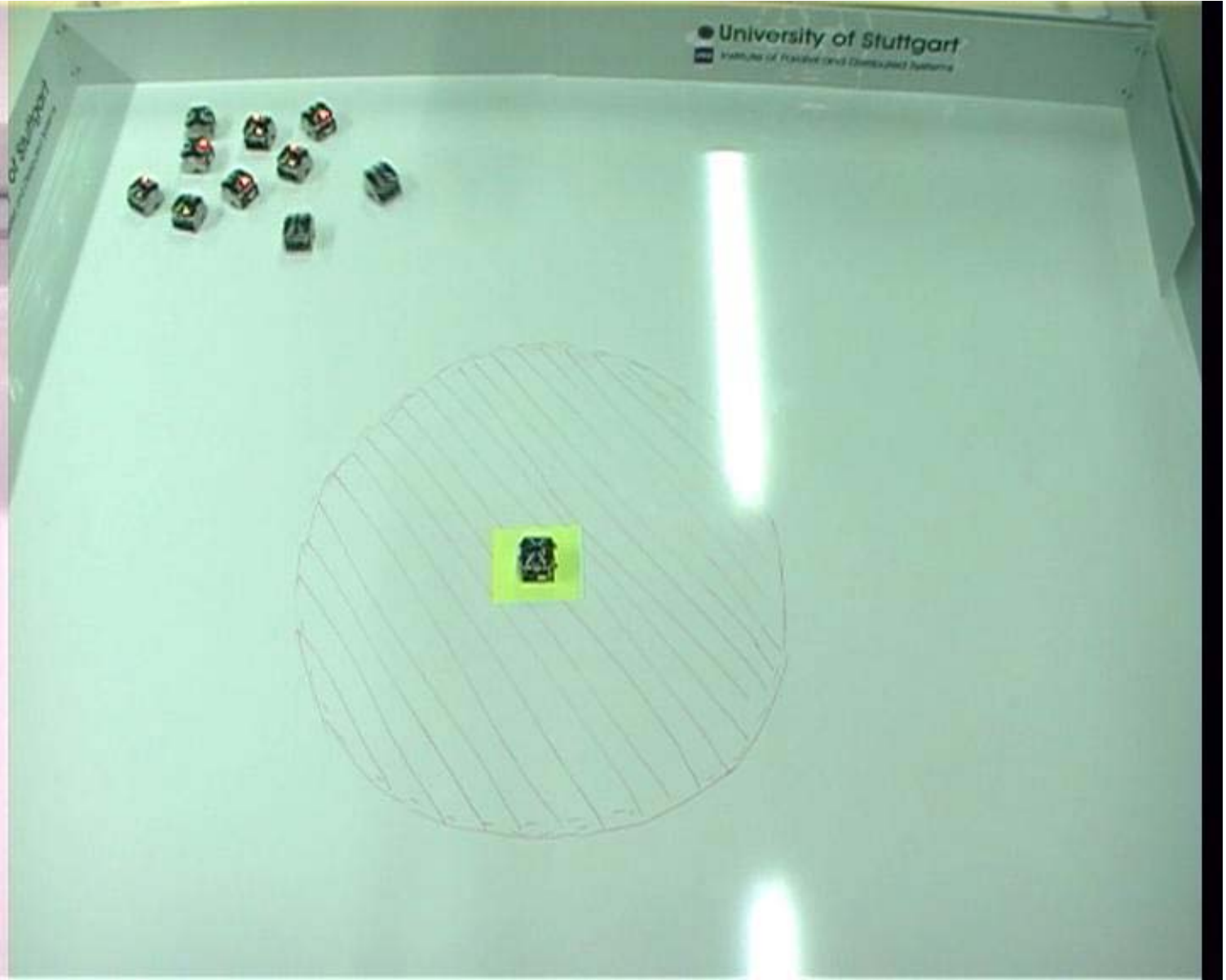}}~
\subfigure[10 sec.]{\includegraphics[width=.32\textwidth]{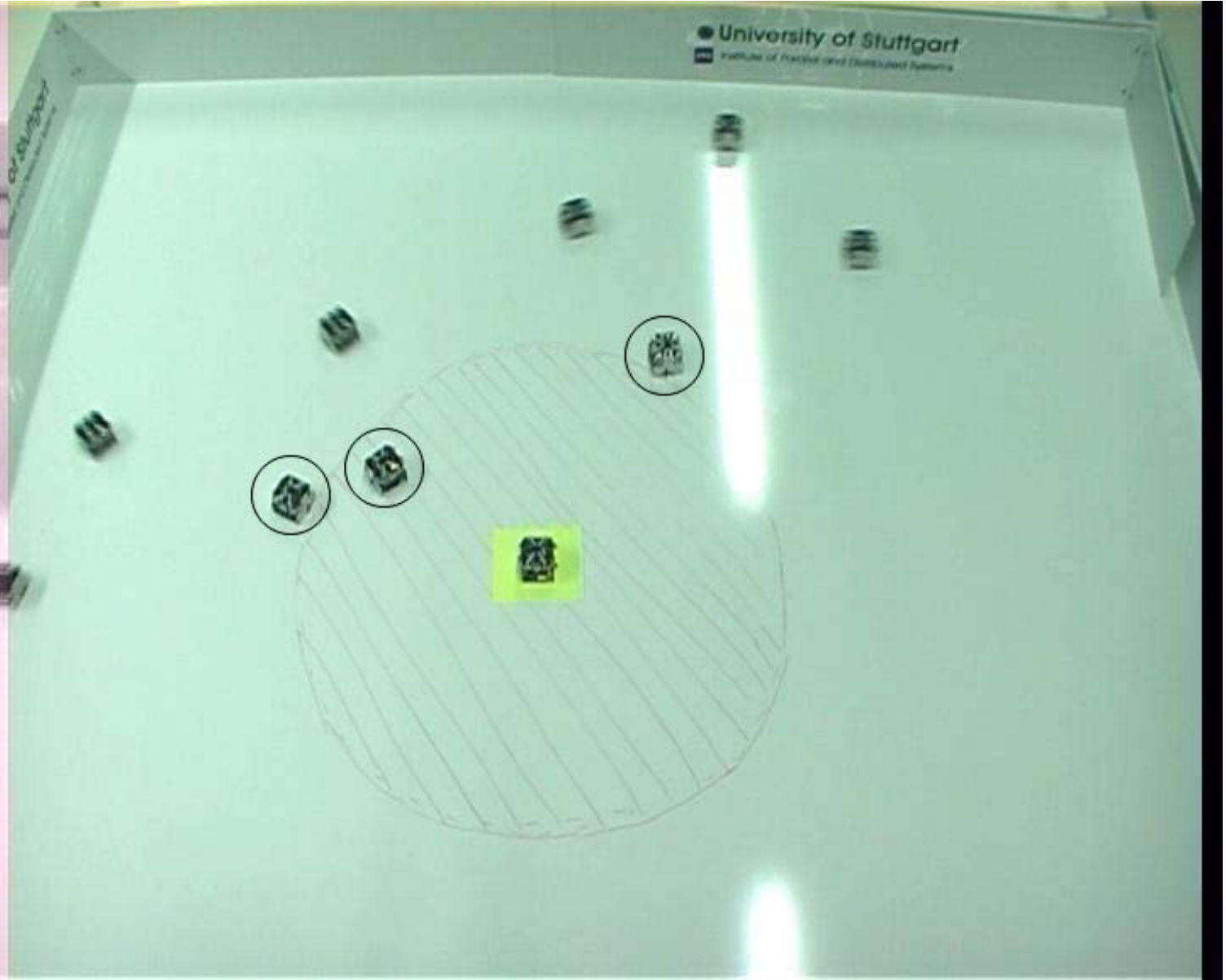}}~
\subfigure[20 sec.]{\includegraphics[width=.32\textwidth]{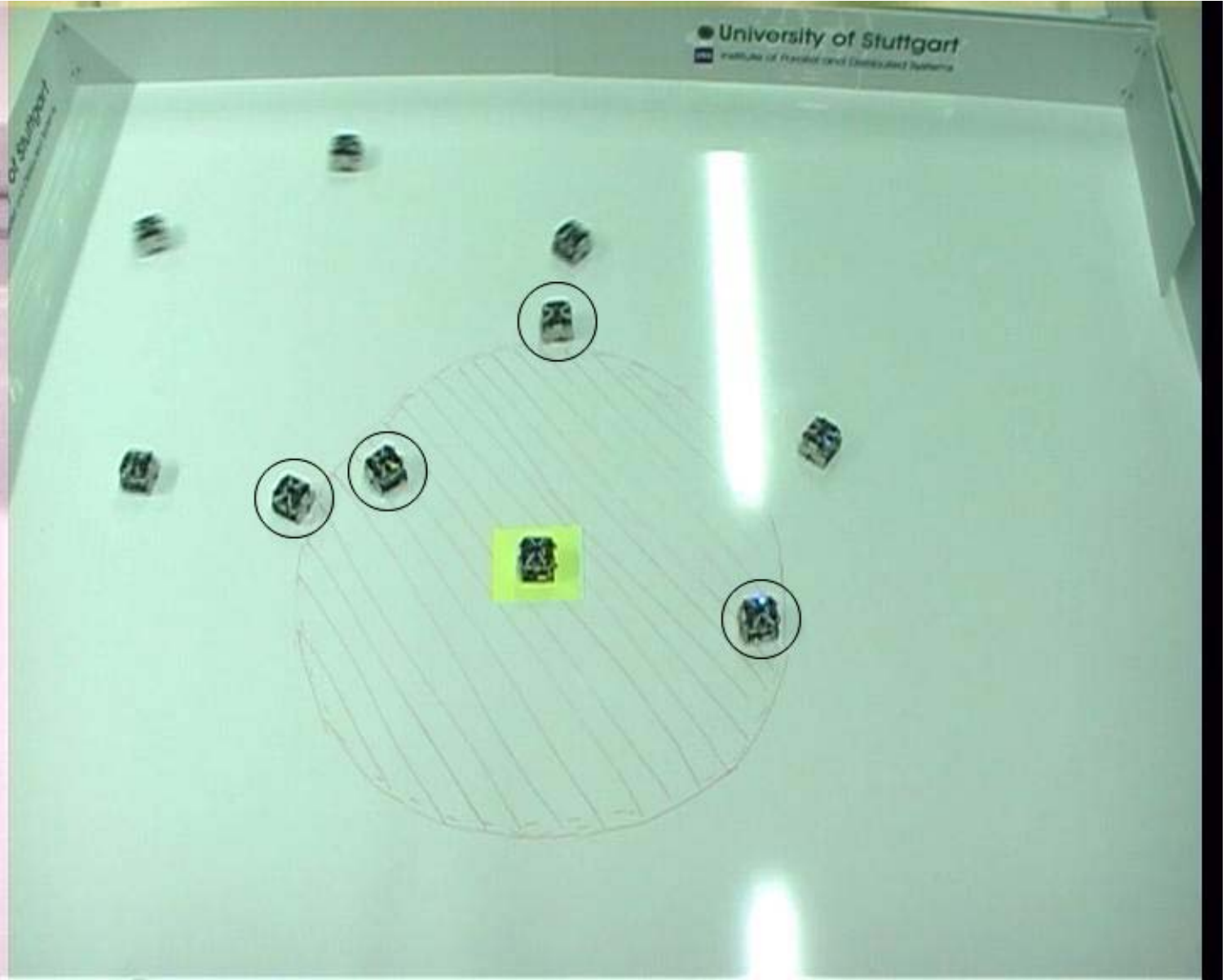}}\\
\subfigure[30 sec.]{\includegraphics[width=.32\textwidth]{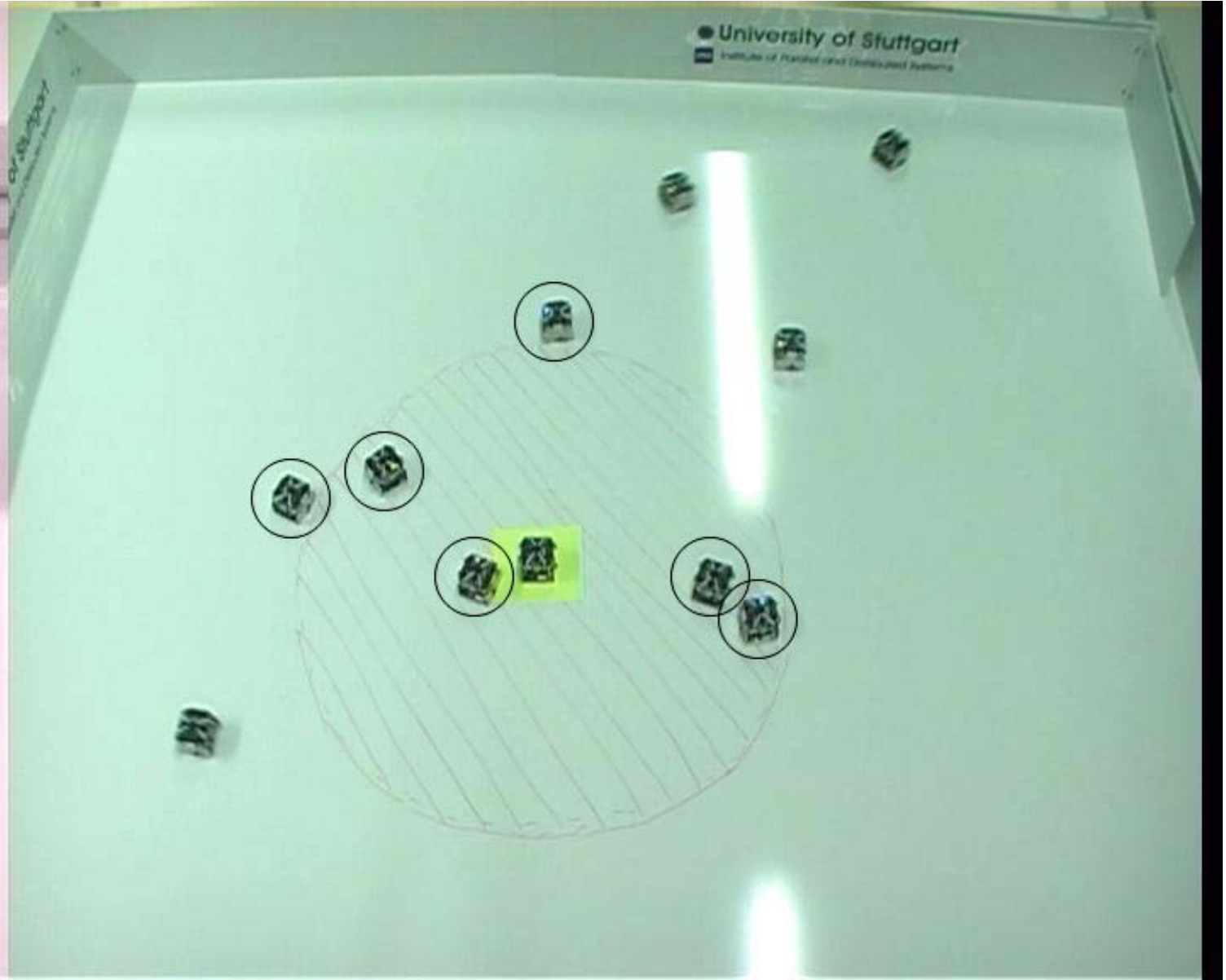}}~
\subfigure[40 sec.]{\includegraphics[width=.32\textwidth]{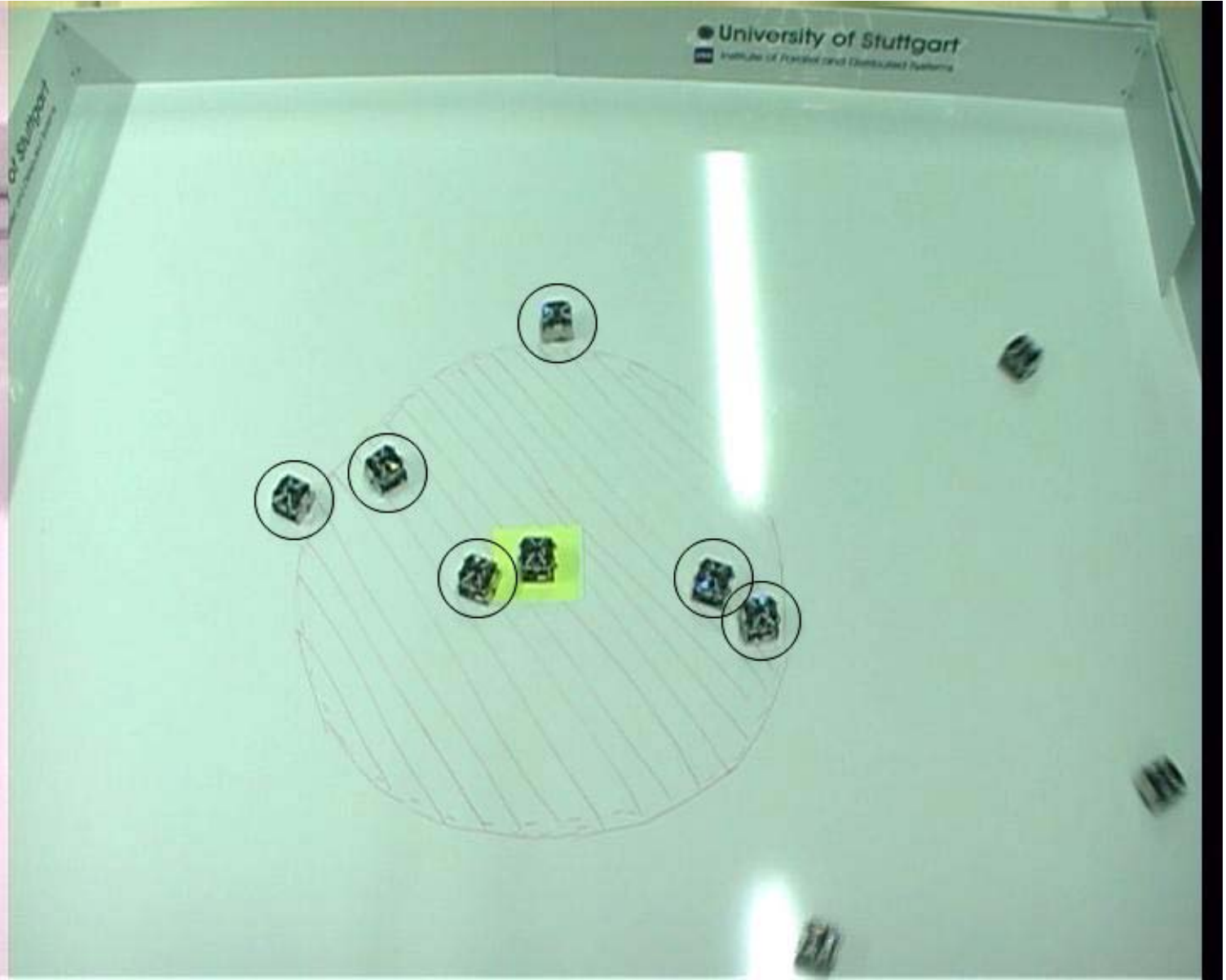}}~
\subfigure[50 sec.]{\includegraphics[width=.32\textwidth]{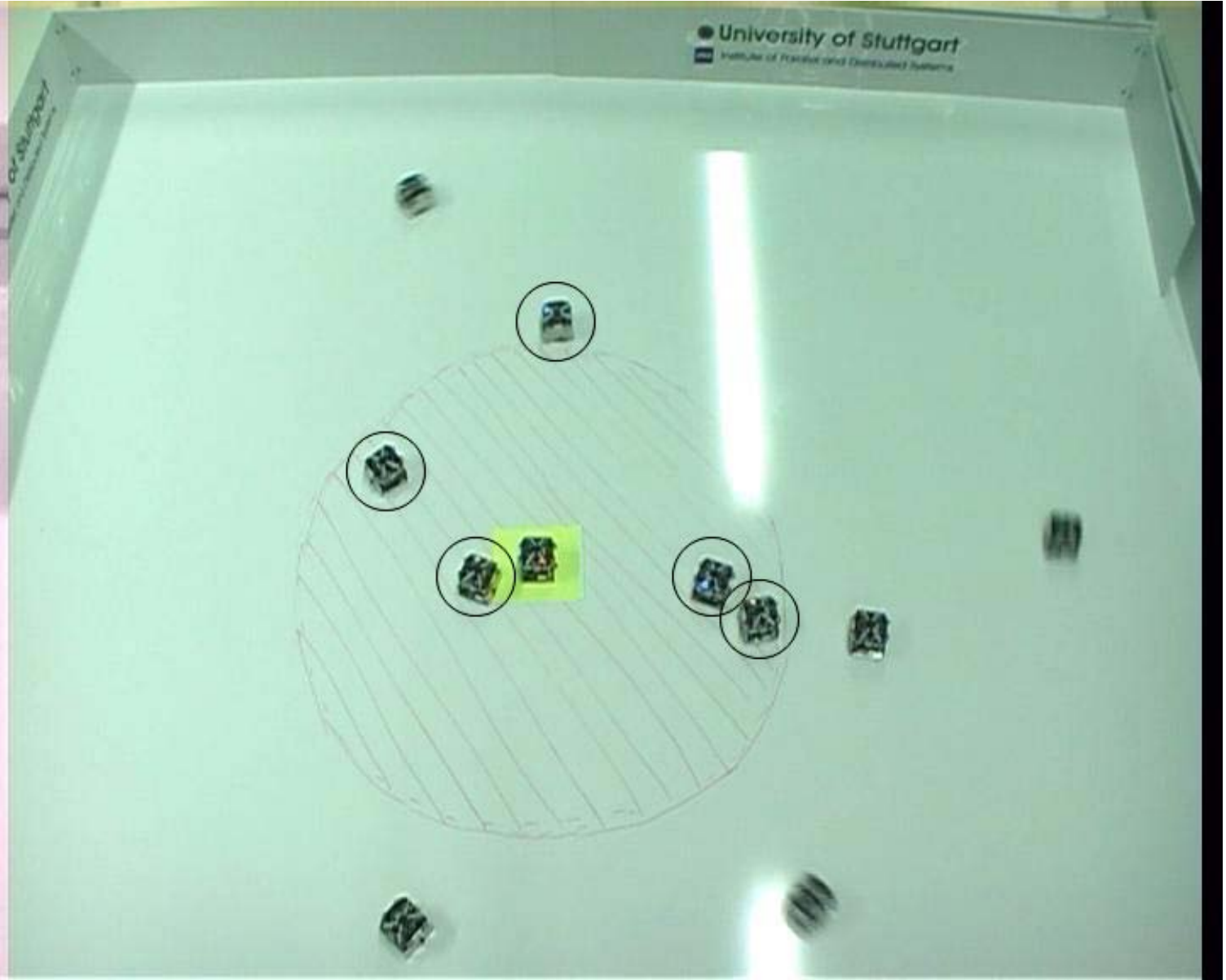}}\\
\subfigure[60 sec.]{\includegraphics[width=.32\textwidth]{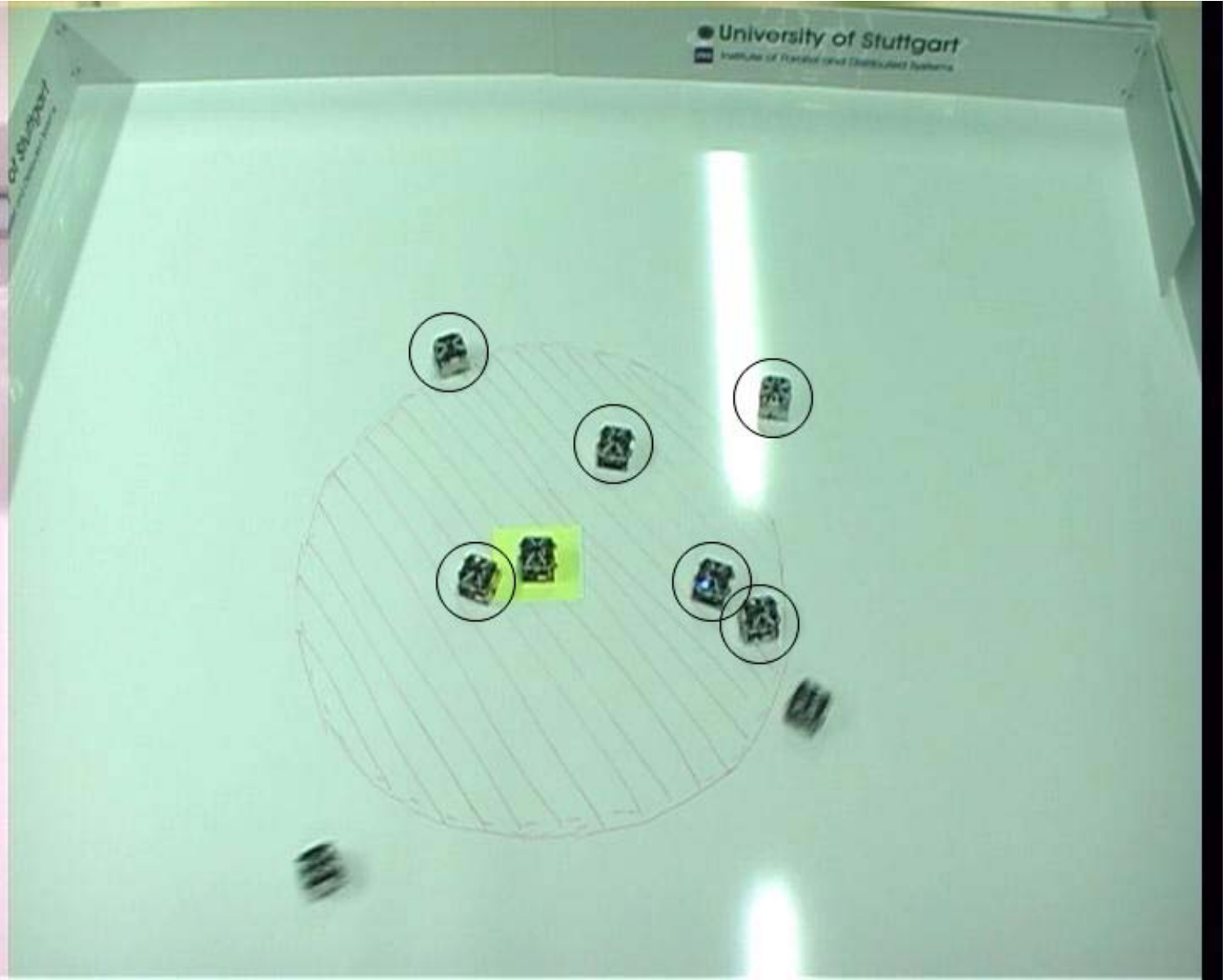}}~
\subfigure[70 sec.]{\includegraphics[width=.32\textwidth]{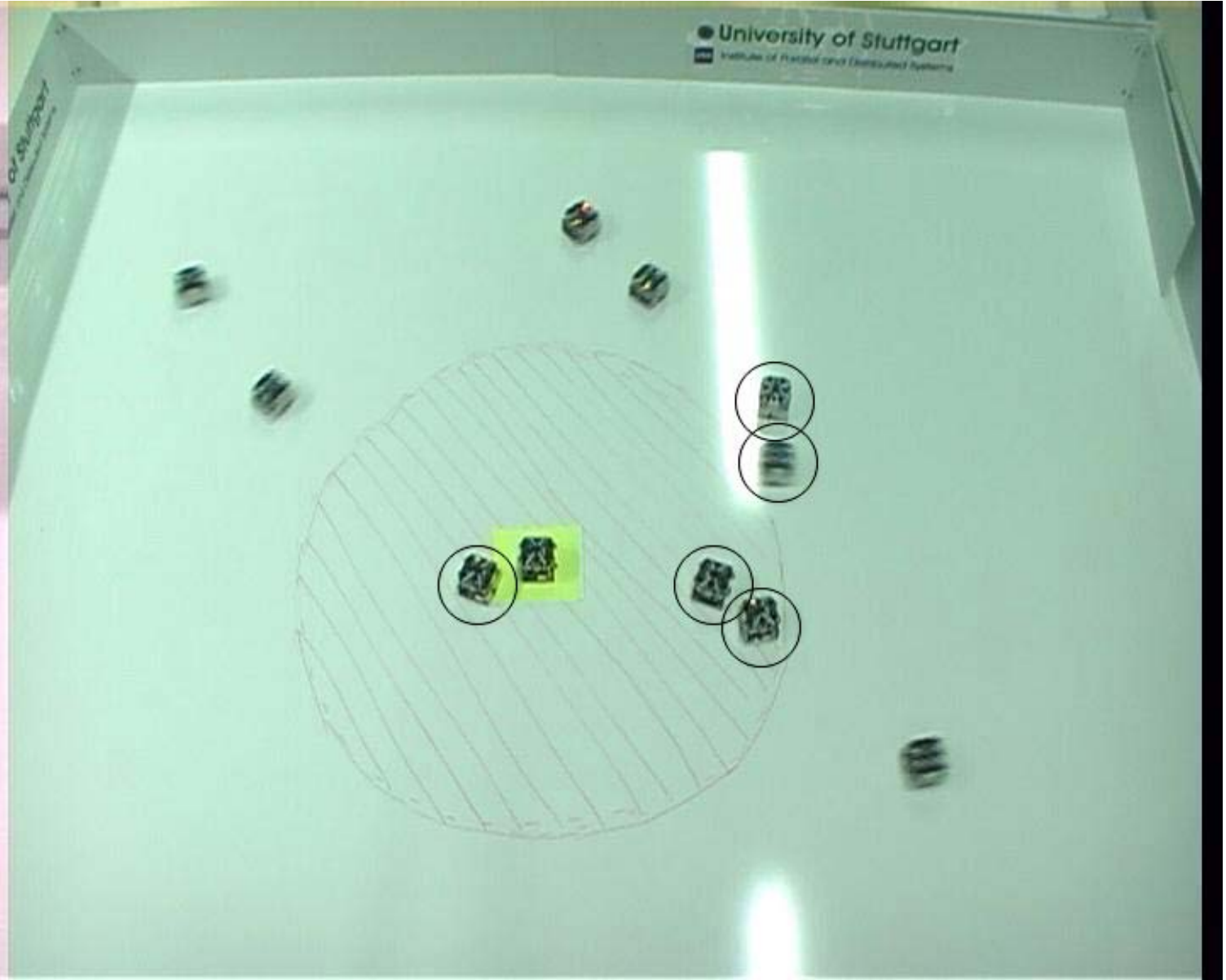}}~
\subfigure[80 sec.]{\includegraphics[width=.32\textwidth]{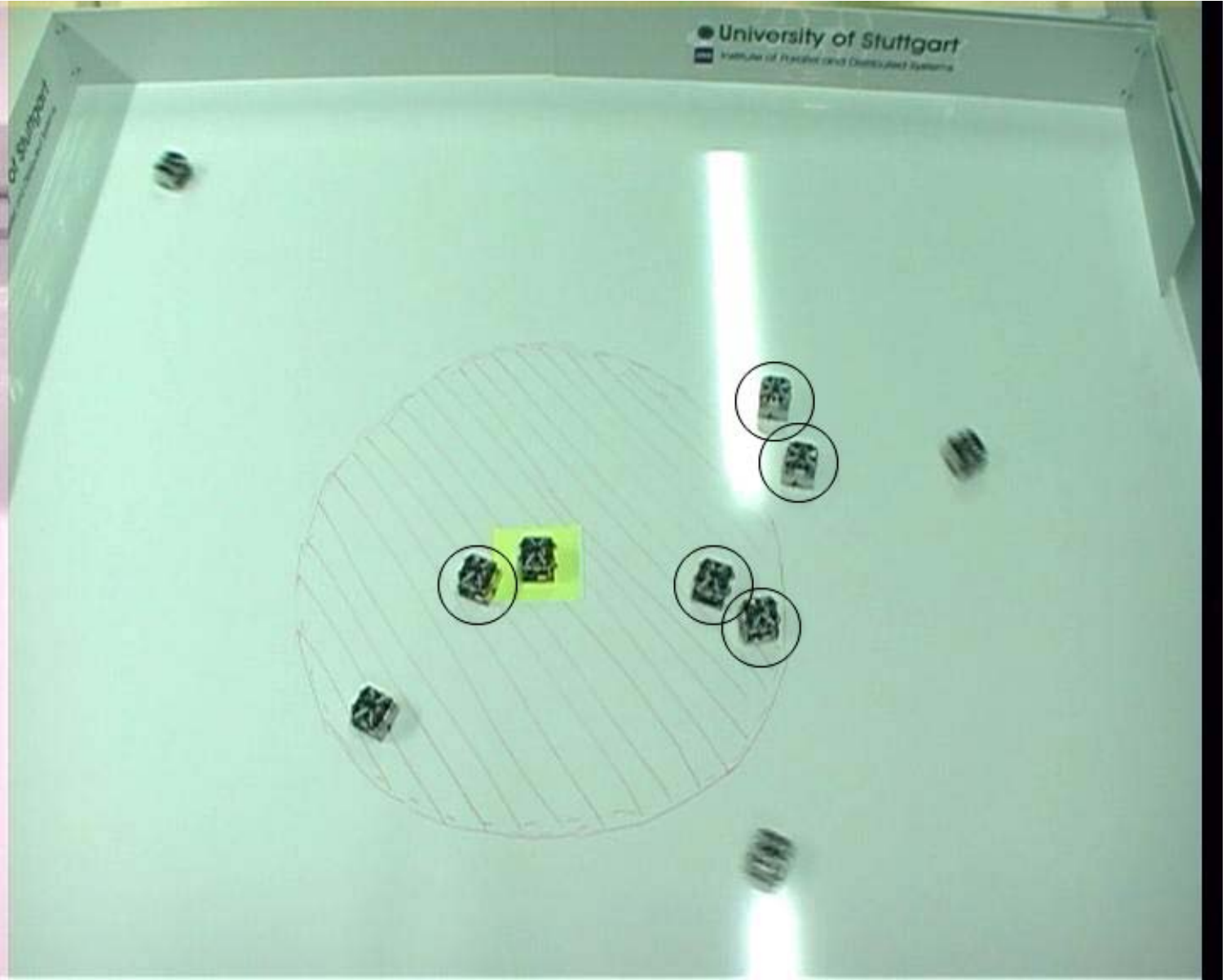}}
\caption{\small Usage of epidemic dynamics for size-dependent clusterization. The robot in the middle of arena is the origin of 1-bit message "I'm here". The marked areal is approximately equal to its communication radius. It is visible that two-robots and three-robots clusters stay longer than separate robots.
\label{fig:clusterization}}
\end{figure}

We put a "seed point" robot in the middle of arena, it plays a role of a landmark, see Fig~\ref{fig:clusterization}. This robot only sends "1", meaning "I'm here". The message is only of a few bits (1 bits when only one robot is a "seed-point" and 4 bits when all 11 robots can create local seed points) so that it is highly efficient in transmitting (or can be even transmitted as analog signal). In contrast to the previous approach, robots cannot change the message, i.e. here we encounter a typical infection dynamics.

Similarly to the previous experiment (this is a standard routine in the Jasmine's software library) all robots, when receiving this message, stop and estimate the signal intensity and receiving channels. In this way, robots know how many other robots are closely to them. This degree of clusterization is a typical temporal information and is used in the feedback. In particular, the waiting time in a cluster is a function of the degree of temporal clusterization~\cite{Kornienko_S06}.

In Fig.~\ref{fig:clusterization} we show the behavior of 11 Jasmine robots within first 80sec. of the experiment, where only one "seed-point" robot is allowed. They build two- and three-robots clusters and these clusters are traveling around the "seed-point" robot. We emphasis that in this approach one bit is sent only by only one robot with very low update rate (one time per 5 seconds). When several seed-points are allowed, other robots can send a reference $k$ (e.g. "2") on the original message ("1").

\section{Conclusion}
\label{sec:conclusion}

In this work we discussed two communication mechanisms, which spread information in a swarm. They are based on package-based approach, however do not use  routing, which is tough for small swarm robots. Instead we utilize virtual field and epidemic algorithms to transmit messages in a system.

The difference between virtual fields and epidemic algorithms consists in creating a spatial gradient and different behavioral procedures and information processing. For instance, virtual fields utilize gradient for navigation purposes, whereas epidemic algorithms focus primarily on population dynamics or temporal effects such as e.g. clusterization. It needs to point out, that the mentioned aspects are related only to the context of robot swarms and not to other applications of virtual fields and epidemic algorithms.

In experiments with different hardware we explore the role of physical embodiment for context-based communication. The selection of the emitting-receiving IR senors has a decisive impact on global communication properties of a swarm. We indicate a possible spectrum of these devices and a selection of particular sensors for the Jasmine robot.

Utilization of spatial information in the case of mobile robots represents some open problems because it is not always possible to establish a relationship between spatial positions of a robot and values of virtual pheromone. Leaving pheromone on immobile objects is of advantage for behavioral algorithms, however is challenging in terms of their practical implementation.

These described mechanisms are not used in separate experiments (at least in those performed in our group), they are always integrated as a basic behavioral-communicating part in more complex scenarios. A long term utilization of these mechanisms indicates their practical relevance and finally leads to a decision to present them as a separate work.


\begin{thebibliography}{10}

\bibitem{Kernbach11-HCR}
Serge Kernbach, editor.
\newblock {\em Handbook of Collective Robotics: Fundamentals and Challenges}.
\newblock Pan Stanford Publishing, Singapore, 2011.

\bibitem{Siciliano08}
Bruno Siciliano and Oussama Khatib, editors.
\newblock {\em Springer Handbook of Robotics}.
\newblock Springer, 2008.

\bibitem{KornienkoS05d}
S.~Kornienko and O.~Kornienko.
\newblock {IR}-based communication and perception in microrobotic swarms.
\newblock In {\em IROS 2005, Edmonton, Canada}, 2005.

\bibitem{Fukuda91}
Toshio Fukuda, Martin Buss, Hidemi Hosokai, and Yoshio Kawauchi.
\newblock Cell structured robotic system cebot: Control, planning and
  communication methods.
\newblock {\em Robotics and Autonomous Systems}, 7(2-3):239--248, 1991.

\bibitem{Liu2010}
Wenguo Liu and Alan F.~T. Winfield.
\newblock Autonomous morphogenesis in self-assembling robots using ir-based
  sensing and local communications.
\newblock In {\em Proceedings of the 7th international conference on Swarm
  intelligence}, ANTS'10, pages 107--118, Berlin, Heidelberg, 2010.
  Springer-Verlag.

\bibitem{Marocco06}
D.~Marocco and S.~Nolfi.
\newblock Origins of communication in evolving robots.
\newblock In Nolfi~S. et~al., editor, {\em SAB06}, pages 789--803. Springer
  Verlag, 2006.

\bibitem{2011arXiv1109.4221K}
S.~{Kernbach}.
\newblock {Three Cases of Connectivity and Global Information Transfer in Robot
  Swarms}.
\newblock {\em ArXiv e-prints}, (2011arXiv1109.4221K), September 2011.

\bibitem{Jimenez05}
M.G. Jim\'enez.
\newblock {\em Cooperative actuation in a large robotic swarm}.
\newblock Master Thesis, University of Stuttgart, Germany, 2006.

\bibitem{Akyildiz04}
Ian~F. Akyildiz, Dario Pompili, and Tommaso Melodia.
\newblock Challenges for efficient communication in underwater acoustic sensor
  networks.
\newblock {\em ACM SIGBED Review}, 2004.

\bibitem{Kornienko_S04}
S.~Kornienko, O.~Kornienko, and P.~Levi.
\newblock Multi-agent repairer of damaged process plans in manufacturing
  environment.
\newblock In {\em Proc. of the 8th Conf. on Intelligent Autonomous Systems
  (IAS-8), Amsterdam, NL}, pages 485--494, 2004.

\bibitem{Kornienko_S05a}
S.~Kornienko, O.~Kornienko, C.~Constantinescu, M.~Pradier, and P.~Levi.
\newblock Cognitive micro-agents: individual and collective perception in
  microrobotic swarm.
\newblock In {\em Proc. of the IJCAI-05 Workshop on Agents in real-time and
  dynamic environments, Edinburgh, UK}, pages 33--42, 2005.

\bibitem{Bonabeau99}
E.~Bonabeau, M.~Dorigo, and G.~Theraulaz.
\newblock {\em Swarm intelligence: from natural to artificial systems}.
\newblock Oxford University Press, New York, 1999.

\bibitem{Kornienko_S05d}
S.~Kornienko, O.~Kornienko, and P.~Levi.
\newblock Minimalistic approach towards communication and perception in
  microrobotic swarms.
\newblock In {\em Proc. of the International Conference on Intelligent Robots
  and Systems (IROS-2005)}, pages 2228--2234, Edmonton, Canada, 2005.

\bibitem{Ebeling98}
W.~Ebeling, J.~Freund, and F.~Schweitzer.
\newblock {\em Komplexe Structuren: Entropie und Information}.
\newblock B.G.~Teubner, 1998.

\bibitem{Kornienko_S05b}
S.~Kornienko, O.~Kornienko, and P.~Levi.
\newblock Collective {AI}: context awareness via communication.
\newblock In {\em Proc. of the IJCAI 2005, Edinburgh, UK}, pages 1464--1470,
  2005.

\bibitem{ma2009}
Z.~Ma and J.~Li, editors.
\newblock {\em Dynamical modeling and analysis of epidemics}.
\newblock World Scientific, 2009.

\bibitem{Kornienko_S05e}
S.~Kornienko, O.~Kornienko, and P.~Levi.
\newblock Swarm embodiment - a new way for deriving emergent behaviour in
  artificial swarms.
\newblock In P.Levi et~al., editor, {\em Autonome Mobile Systeme (AMS'05)},
  pages 25--32, 2005.

\bibitem{Kornienko_OS01}
O.~Kornienko, S.~Kornienko, and P.~Levi.
\newblock Collective decision making using natural self-organization in
  distributed systems.
\newblock In {\em Proc. of Int. Conf. on Computational Intelligence for
  Modelling, Control and Automation (CIMCA'2001), Las Vegas, USA}, pages
  460--471, 2001.

\bibitem{Zhang1098}
Haibo Zhang and Hong Shen.
\newblock Energy-efficient beaconless geographic routing in wireless sensor
  networks.
\newblock {\em IEEE Transactions on Parallel and Distributed Systems},
  21:881--896, 2010.

\bibitem{Payton01}
D.W. Payton, M.~Daily, R.~Estowski, M.~Howard, and C.~Lee.
\newblock Pheromone robotic.
\newblock {\em Auton. Robots}, 11(3):319--324, 2001.

\bibitem{Kornienko_S04b}
S.~Kornienko, O.~Kornienko, and P.~Levi.
\newblock About nature of emergent behavior in micro-systems.
\newblock In {\em Proc. of the Int. Conf. on Informatics in Control, Automation
  and Robotics (ICINCO 2004), Setubal, Portugal}, pages 33--40, 2004.

\bibitem{Kornienko_S04a}
S.~Kornienko, O.~Kornienko, and P.~Levi.
\newblock Generation of desired emergent behavior in swarm of micro-robots.
\newblock In R.~{Lopez de Mantaras} and L.~Saitta, editors, {\em Proc. of the
  16th European conference on artificial intelligence (ECAI 2004), Valencia,
  Spain}, pages 239--243. Amsterdam: IOS Press, 2004.

\bibitem{Kernbach08online}
L.~K{\"o}nig, K.~Jebens, Serge Kernbach, and Paul Levi.
\newblock Stability of on-line and on-board evolving of adaptive collective
  behavior.
\newblock In Herman Bruyninckx, Libor Preucil, and Miroslav Kulich, editors,
  {\em European Robotics Symposium 2008}, pages 293--302. 2008.

\bibitem{Kernbach08Permis}
Serge Kernbach, Eugen Meister, Florian Schlachter, Kristof Jebens, Marc
  Szymanski, Jens Liedke, Davide Laneri, Lutz Winkler, Thomas Schmickl, Ronald
  Thenius, Paolo Corradi, and Leonardo Ricotti.
\newblock Symbiotic robot organisms: {REPLICATOR} and {SYMBRION} projects.
\newblock In {\em Proceedings of the 8th Workshop on Performance Metrics for
  Intelligent Systems}, PerMIS '08, pages 62--69, New York, NY, USA, 2008. ACM.

\bibitem{Nembrini02}
J.~Nembrini, A.~Winfield, and C.~Melhuish.
\newblock Minimalist coherent swarming of wireless connected autonomous mobile
  robots.
\newblock In {\em Proc. of Int. Conf. on Simulation of Artificial Behaviour},
  Edinburgh, 2002.

\bibitem{kernbach09adaptive}
Serge Kernbach, Heiko Hamann, J\"{u}rgen Stradner, Ronald Thenius, Thomas
  Schmickl, Karl Crailsheim, A.~C.~van Rossum, Michele Sebag, Nicolas Bredeche,
  Yao Yao, Guy Baele, Yves Van~de Peer, Jon Timmis, Maizura Mohktar, Andy
  Tyrrell, A.~E. Eiben, S.~P. McKibbin, Wenguo Liu, and Alan F.~T. Winfield.
\newblock On adaptive self-organization in artificial robot organisms.
\newblock In {\em Proceedings of the 2009 Computation World: Future Computing,
  Service Computation, Cognitive, Adaptive, Content, Patterns},
  COMPUTATIONWORLD '09, pages 33--43, Washington, DC, USA, 2009. IEEE Computer
  Society.

\bibitem{Kornienko_S03A}
S.~Kornienko, O.~Kornienko, and P.~Levi.
\newblock Flexible manufacturing process planning based on the multi-agent
  technology.
\newblock In {\em Proc. of the 21st IASTED Int. Conf. on AI and Applications
  (AIA '2003), Innsbruck, Austria}, pages 156--161, 2003.

\bibitem{Geider06}
R.~Geider.
\newblock {\em Development of context-based communication protocols for the
  microrobot 'Jasmine'}.
\newblock Studienarbeit, University of Stuttgart, Germany, 2006.

\bibitem{Lloyd1996}
Alun~L. Lloyd and Robert~M. May.
\newblock Spatial heterogeneity in epidemic models.
\newblock {\em Journal of Theoretical Biology}, 179(1):1 -- 11, 1996.

\bibitem{Chaintreau:2009}
Augustin Chaintreau, Jean-Yves Le~Boudec, and Nikodin Ristanovic.
\newblock The age of gossip: spatial mean field regime.
\newblock In {\em Proceedings of the eleventh international joint conference on
  Measurement and modeling of computer systems}, SIGMETRICS '09, pages
  109--120, New York, NY, USA, 2009. ACM.

\bibitem{Kernbach09Platform}
S.~Kernbach, E.~Meister, O.~Scholz, R.~Humza, J.~Liedke, L.~Ricotti, J.~Jemai,
  J.~Havlik, and W.~Liu.
\newblock Evolutionary robotics: The next-generation-platform for on-line and
  on-board artificial evolution.
\newblock In {\em IEEE Congress on Evolutionary Computation, CEC '09}, pages
  1079 --1086, 2009.

\bibitem{Kornienko_S06}
S.~Kernbach, R.~Thenius, O.~Kernbach, and T.~Schmickl.
\newblock Re-embodiment of honeybee aggregation behavior in artificial
  micro-robotic system.
\newblock {\em Adaptive Behavior}, 17(3):237--259, 2009.

\bibitem{Kernbach08_2}
S.~Kernbach, L.~Ricotti, J.~Liedke, P.~Corradi, and M.~Rothermel.
\newblock Study of macroscopic morphological features of symbiotic robotic
  organisms.
\newblock In {\em Proceedings of the workshop on self-reconfigurable robots,
  IROS08, Nice}, pages 18--25, 2008.

\bibitem{Suzuki95}
S.~Suzuki, H.~Asama, A.~Uegaki, S.~Kotosaka, T.~Fujita, A.~Matsumoto,
  H.~Kaetsu, and I.~Endo.
\newblock An infra-red sensory system with local communication for cooperative
  multiple mobile robots.
\newblock In {\em Proc. of International Conference on Intelligent Robots},
  pages 220--225, 1995.

\bibitem{Levi10}
Paul Levi and Serge Kernbach, editors.
\newblock {\em Symbiotic Multi-Robot Organisms: Reliability, Adaptability,
  Evolution}.
\newblock Springer Verlag, 2010.

\bibitem{Kernbach08}
S.~Kernbach.
\newblock {\em Structural Self-organization in Multi-Agents and Multi-Robotic
  Systems}.
\newblock Logos Verlag, Berlin, 2008.

\bibitem{Levi99}
P.~Levi, M.~Schanz, S.~Kornienko, and O.~Kornienko.
\newblock Application of order parameter equation for the analysis and the
  control of nonlinear time discrete dynamical systems.
\newblock {\em Int. J. Bifurcation and Chaos}, 9(8):1619--1634, 1999.

\end{thebibliography}

\end{document}